\documentclass[11pt, a4paper]{lumia}
\pdfoutput=1

\usepackage[sort&compress]{natbib}
\usepackage{xspace}

\theoremstyle{plain}
\newtheorem{theorem}{Theorem}

\newtheorem*{proposition*}{Proposition}

\theoremstyle{definition}

\theoremstyle{definition}

\def\eqref#1{equation~\ref{#1}}

\usepackage{graphicx}           
\usepackage{tikz}               
\usepackage[edges]{forest}      

\usepackage{url}                
\usepackage{xurl}               

\usepackage{array}              
\usepackage{longtable}          
\usepackage{multirow}           
\usepackage{makecell}           
\usepackage{ragged2e}           

\usepackage{mathtools}          
\usepackage{nicefrac}           

\usepackage{algorithm}          
\usepackage{algorithmicx}       
\usepackage{algpseudocode}      
\usepackage{listings}           

\usepackage{subcaption}         
\usepackage{wrapfig}            
\usepackage[export]{adjustbox}  

\usepackage{changes}            
\usepackage{xspace}             
\usepackage[normalem]{ulem}     
\usepackage{CJKutf8}            

\usepackage[tikz]{bclogo}       
\usepackage[framemethod=tikz]{mdframed} 

\usepackage{placeins}
\usepackage{cleveref}
\usepackage{lipsum}             
\usepackage{tocloft}            
\usepackage{afterpage}          
\usepackage{bbding}             
\usepackage{epigraph}           
\usepackage{minitoc}            
\usepackage{multicol}           
\usepackage{textgreek}          

\newcommand{\hlfirst}[1]{\begingroup\setlength{\fboxsep}{1pt}\colorbox[HTML]{CFE2FF}{#1}\endgroup}
\newcommand{\hlsecond}[1]{\begingroup\setlength{\fboxsep}{1pt}\colorbox[HTML]{E8F1FF}{#1}\endgroup}

\definecolor{mygreen}{rgb}{0.29, 0.7, 0.48}
\definecolor{darksalmon}{rgb}{0.91, 0.59, 0.48}
\definecolor{mygrey}{gray}{0.4}

\usepackage[most]{tcolorbox}
\usepackage{makecell}
\newcolumntype{P}[1]{>{\RaggedRight\arraybackslash}p{#1}}

\definecolor{uclablue}{RGB}{39, 116, 174}
\definecolor{bigaired}{RGB}{156, 0, 0}
\definecolor{myblue}{HTML}{598BE7}
\definecolor{mildblue}{RGB}{31,119,180}
\definecolor{sectionblue}{RGB}{70, 130, 180}
\definecolor{methodblue}{RGB}{0, 150, 136}
\definecolor{bgblue}{RGB}{245,243,253}
\definecolor{ttblue}{RGB}{91,194,224}
\definecolor{mygreen}{rgb}{0.64, 0.56, 0.88}
\definecolor{myyellow}{rgb}{0.68, 0.6, 0.1}
\definecolor{fancygreen}{rgb}{0.33, 0.68, 0.20}
\definecolor{salmon}{rgb}{0.94, 0.52, 0.49}
\definecolor{tablegreen}{rgb}{0.82, 0.94, 0.75}
\definecolor{tableblue}{rgb}{0.81, 0.90, 0.94}
\definecolor{tablered}{rgb}{0.97, 0.85, 0.85}
\definecolor{tableorange}{rgb}{0.96, 0.85, 0.81}
\definecolor{myorange}{rgb}{1.0, 0.49, 0.0}
\definecolor{tlgreen}{rgb}{0.33, 0.68, 0.20}
\definecolor{darkgreen}{RGB}{0,100,0}
\definecolor{darkred}{RGB}{200, 0, 0}
\definecolor{customyellow}{HTML}{FFFACD}
\definecolor{refinegreen}{RGB}{0, 128, 75}
\definecolor{scoregreen}{RGB}{34, 139, 34}
\definecolor{hidden-blue}{RGB}{194,232,247}
\definecolor{hidden-black}{RGB}{20,68,106}
\definecolor{yes}{HTML}{C6EFCE}
\definecolor{no}{HTML}{FFC7CE}
\definecolor{partial}{HTML}{FFEB9C}
\definecolor{external}{HTML}{D9E1F2}
\definecolor{hdr}{HTML}{F2F2F2}
\definecolor{GRPOrow}{gray}{0.96}
\definecolor{FlowRLrow}{RGB}{225,236,255}
\definecolor{FlowBlue}{RGB}{80,120,210}
\definecolor{GRPOGray}{gray}{0.35}

\setlist[itemize]{leftmargin=20pt, noitemsep, topsep=0pt}

\newcommand{\cmark}{\textcolor{darkgreen}{\boldmath$\checkmark$}}
\newcommand{\xmark}{\textcolor{darkred}{\boldmath$\times$}}

\newenvironment{itemize*}%
 {\leftmargini=10pt\begin{itemize}%
  \setlength{\itemsep}{0pt}%
  \setlength{\parskip}{0pt}%
  }%
 {\end{itemize}}

\newenvironment{enumerate*}%
 {\begin{enumerate}%
  \setlength{\itemsep}{0pt}%
  \setlength{\parskip}{0pt}}%
 {\end{enumerate}}

\newcommand{\cellstatus}[1]{%
  \begingroup
  \StrTrim{#1}[\statusval]%
  \IfStrEq{\statusval}{Yes}{\cellcolor{yes}\cmark}{}%
  \IfStrEq{\statusval}{No}{\cellcolor{no}\xmark}{}%
  \IfBeginWith{\statusval}{Yes (}{\cellcolor{yes}\cmark~\textit{\statusval\unskip}}{}%
  \IfStrEq{\statusval}{Partial}{\cellcolor{partial}\textbf{Partial}}{}%
  \IfStrEq{\statusval}{External}{\cellcolor{external}\textbf{External}}{}%
  \endgroup
}

\newtcolorbox{myboxi}[1][]{
  breakable,
  title=#1,
  colback=red!5,
  colbacktitle=red!5,
  coltitle=black,
  fonttitle=\bfseries,
  bottomrule=0pt,
  toprule=0pt,
  leftrule=2pt,
  rightrule=2pt,
  titlerule=0pt,
  arc=0pt,
  outer arc=0pt,
  colframe=red,
}

\newtcolorbox{myboxnote}[1][]{
  breakable,
  title=#1,
  colback=orange!0,
  colbacktitle=orange!0,
  coltitle=black,
  fonttitle=\bfseries,
  bottomrule=0pt,
  toprule=0pt,
  leftrule=2pt,
  rightrule=2pt,
  titlerule=0pt,
  arc=0pt,
  outer arc=0pt,
  colframe=orange,
}

\newtcolorbox{myboxii}[1][]{
  breakable,
  freelance,
  title=#1,
  colback=white,
  colbacktitle=white,
  coltitle=black,
  fonttitle=\bfseries,
  bottomrule=0pt,
  boxrule=0pt,
  colframe=white,
  overlay unbroken and first={
  \draw[red!75!black,line width=3pt]
    ([xshift=5pt]frame.north west) -- 
    (frame.north west) -- 
    (frame.south west);
  \draw[red!75!black,line width=3pt]
    ([xshift=-5pt]frame.north east) -- 
    (frame.north east) -- 
    (frame.south east);
  },
  overlay unbroken app={
  \draw[red!75!black,line width=3pt,line cap=rect]
    (frame.south west) -- 
    ([xshift=5pt]frame.south west);
  \draw[red!75!black,line width=3pt,line cap=rect]
    (frame.south east) -- 
    ([xshift=-5pt]frame.south east);
  },
  overlay middle and last={
  \draw[red!75!black,line width=3pt]
    (frame.north west) -- 
    (frame.south west);
  \draw[red!75!black,line width=3pt]
    (frame.north east) -- 
    (frame.south east);
  },
  overlay last app={
  \draw[red!75!black,line width=3pt,line cap=rect]
    (frame.south west) --
    ([xshift=5pt]frame.south west);
  \draw[red!75!black,line width=3pt,line cap=rect]
    (frame.south east) --
    ([xshift=-5pt]frame.south east);
  },
}

\tcbset{
  takeawaysbox/.style={
    title=Takeaways,
    colback=lightblue!80,
    colframe=black,
    fonttitle=\bfseries\small,
    coltitle=white,
    colbacktitle=black,
    enhanced,
    attach boxed title to top left={xshift=2.5mm,yshift=-2.5mm},
    boxed title style={rounded corners, size=small, colframe=black, colback=black},
    width=\linewidth,
    arc=3.5mm
  }
}

\mdfdefinestyle{mystyle}{%
  rightline=true,
  innerleftmargin=10,
  innerrightmargin=10,
  outerlinewidth=3pt,
  topline=false,
  rightline=true,
  bottomline=false,
  skipabove=\topsep,
  skipbelow=\topsep
}

\tikzset{%
    every node/.style={font=\tiny},
    parent/.style =          {align=center,text width=2cm,rounded corners=3pt, line width=0.3mm, fill=gray!10,draw=gray!80},
    child/.style =           {align=center,text width=2.0cm,rounded corners=3pt, fill=blue!10,draw=blue!80,line width=0.3mm},
    grandchild/.style =      {align=center,text width=2cm,rounded corners=3pt},
    greatgrandchild/.style = {align=center,text width=1.5cm,rounded corners=3pt},
    greatgrandchild2/.style = {align=center,text width=1.5cm,rounded corners=3pt},    
    referenceblock/.style =  {align=center,text width=1.5cm,rounded corners=2pt},
    pretrain/.style =           {align=center,text width=2.0cm,rounded corners=3pt, fill=blue!10,draw=blue!80,line width=0.3mm},   
    pretrain_work/.style =           {align=center, text width=8.5cm,rounded corners=3pt, fill=blue!10,draw=blue!0,line width=0.3mm},  
    template/.style =           {align=center,text width=2.0cm,rounded corners=3pt, fill=red!10,draw=red!80,line width=0.3mm},   
    template_work/.style =           {align=center,text width=8.5cm,rounded corners=3pt, fill=red!10,draw=red!0,line width=0.3mm},    
    answer/.style =           {align=center,text width=2.0cm,rounded corners=3pt, fill= cyan!10,draw= cyan!80,line width=0.3mm},   
    answer_work/.style =           {align=center,text width=8.5cm,rounded corners=3pt, fill= cyan!10,draw= cyan!0,line width=0.3mm},      
    multiple/.style =           {align=center,text width=2.0cm,rounded corners=3pt, fill= orange!10,draw= orange!80,line width=0.3mm},   
    multiple_work/.style =           {align=center,text width=8.5cm,rounded corners=3pt, fill= orange!10,draw= orange!0,line width=0.3mm},        
    tuning/.style =           {align=center,text width=2.0cm,rounded corners=3pt, fill= magenta!10,draw= magenta!80,line width=0.3mm},   
    tuning_work/.style =           {align=center,text width=8.5cm,rounded corners=3pt, fill= magenta!10,draw= magenta!0,line width=0.3mm},          
}

\newcommand{\lstbg}[3][0pt]{{\fboxsep#1\colorbox{#2}{\strut #3}}}

\lstdefinelanguage{diff}{
  basicstyle=\ttfamily\small,
  morecomment=[f][\lstbg{red!20}]-,
  morecomment=[f][\lstbg{green!20}]+,
}

\lstdefinelanguage{diffpython}{
  language=diff,
  morekeywords={def, if, else, for, while, return, import, from, as, class, with, try, except, finally, raise, lambda, and, or, not, in, is, None, True, False},
  morecomment=[l]{\#},
  morestring=[b]",
  morestring=[b]',
}

\usepackage{dsfont}
\usepackage{fvextra}
\usepackage{fancyvrb}

\setheadertext{
  \includegraphics[height=1.2cm]{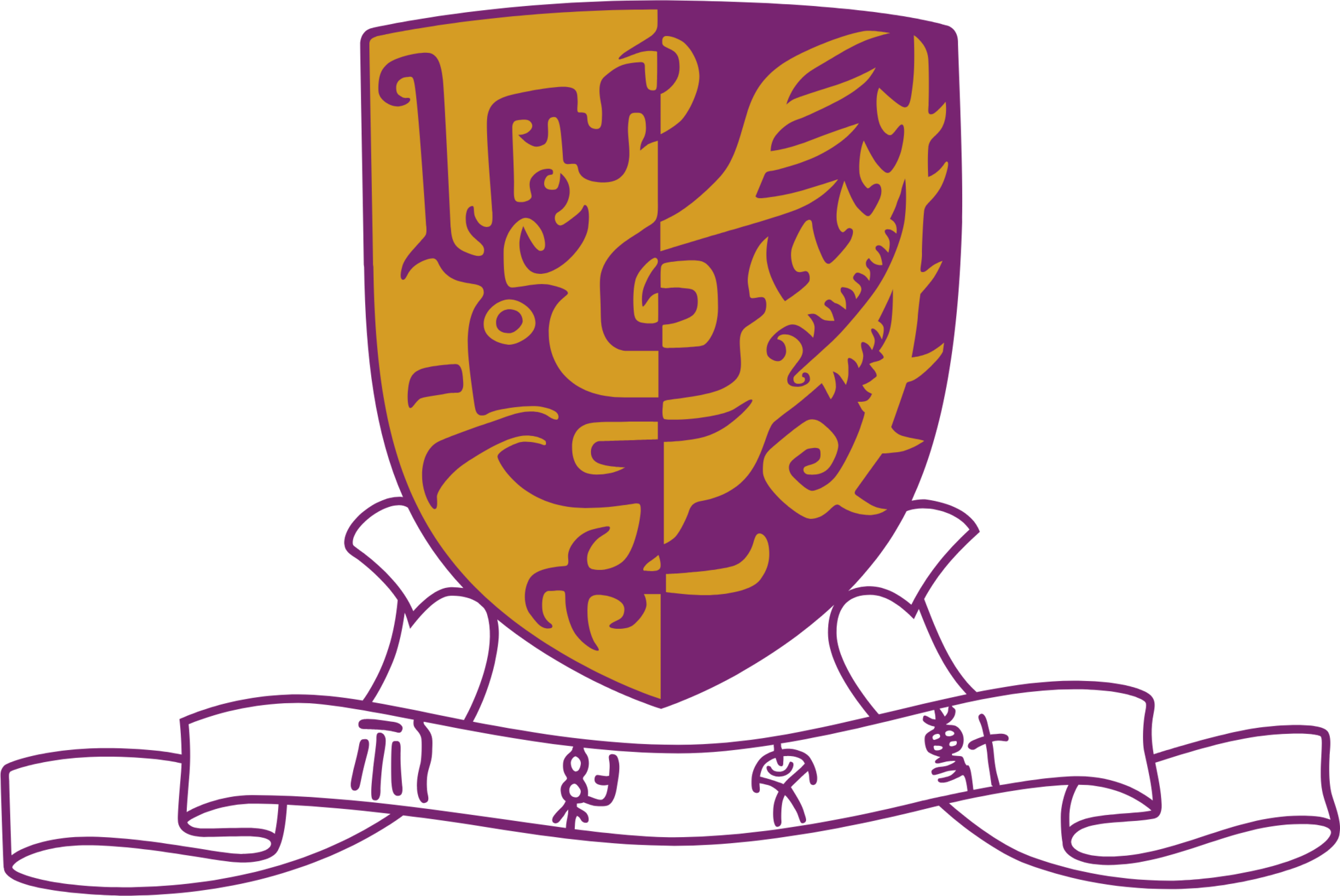} \hspace{0.1cm}%
  \includegraphics[height=1.2cm]{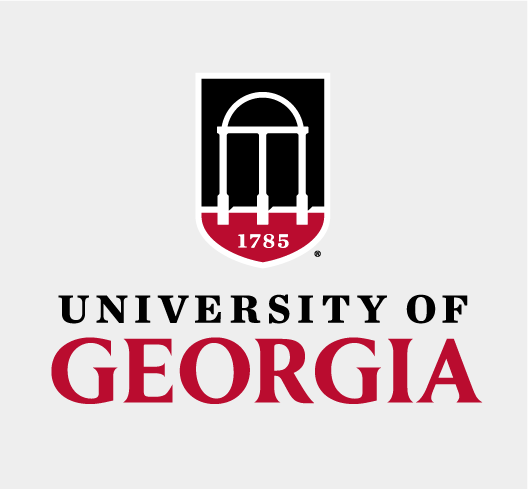} \hspace{0.1cm}%
  \includegraphics[height=1.2cm]{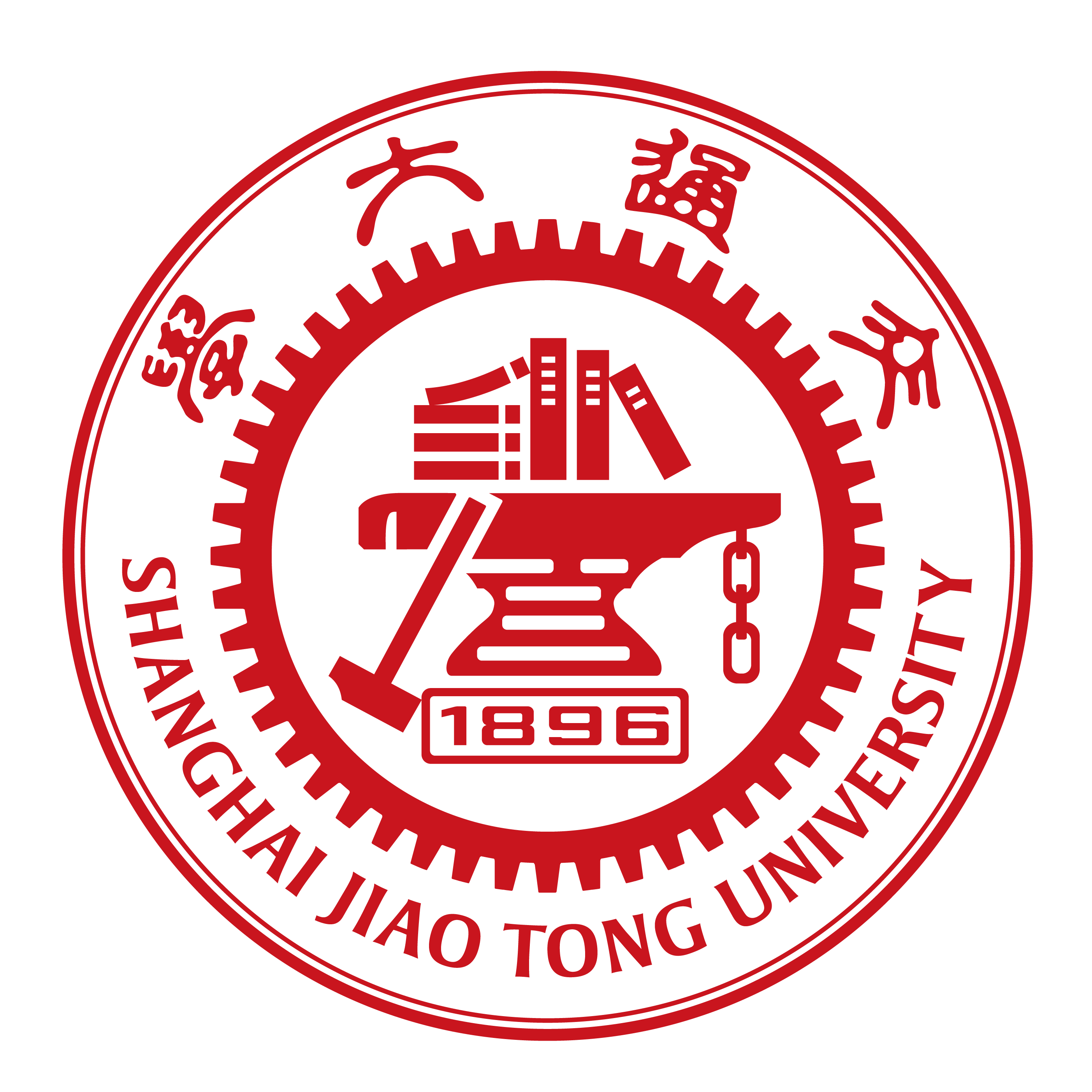} \hspace{0.1cm}%
  \includegraphics[height=1.2cm]{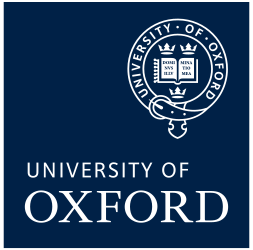}
}

\correspondingemail{
  {\emailicon\ }\href{mailto:xxythecoder@gmail.com}{xxythecoder@gmail.com} \quad $^\ast$Equal contribution. \quad $^\dagger$Corresponding authors.
}

\githublink{https://github.com/xxyQwQ/SuffiCE}

\huggingfacelink{https://huggingface.co/collections/xxyQwQ/suffice}

\title{SuffiCE: Densifying Supervision for Context Compaction in Agentic Reinforcement Learning with Suffix-Compatibility Estimation}
\setheadertitle{SuffiCE: Densifying Supervision for Context Compaction in Agentic RL with Suffix-Compatibility Estimation}

\author{
  Xiangyuan Xue$^{1\ast}$, Yifan Zhou$^{2\ast}$, Shengji Tang$^{1}$, Zidong Wang$^{1}$, Zeyu Lu$^{3}$, Zhenfei Yin$^{4\dagger}$, Wanli Ouyang$^{1\dagger}$ \\
  $^1$The Chinese University of Hong Kong\quad $^2$University of Georgia\quad $^3$Shanghai Jiao Tong University\quad $^4$University of Oxford
}

\begin{document}

\begin{abstract}
Integrating context compaction techniques into agentic reinforcement learning (RL) allows large-language-model (LLM)-based agents to overcome the limit of the context window and to explore and learn more thoroughly on long-horizon tasks. However, current compaction-integrated agentic RL supervises trajectories only with sparse outcome rewards, which cannot tell whether a failed task results from errors or omissions in a summary or from incorrect decisions. This difficulty in credit assignment leads to a cold-start problem, making it hard for the model to learn to produce good summaries from scratch. In this work, we propose \textbf{Suffi}x-\textbf{C}ompatibility \textbf{E}stimation (SuffiCE). SuffiCE uses deferred summarization to postpone the application of a summary and obtain reference actions under the original context, and then uses compatibility scoring to compare the likelihoods of these actions under the original and compacted contexts, which provides a local reward for the summary without additional supervision. Experiments across various benchmarks and backbones show that SuffiCE outperforms all baselines, overcoming the cold-start problem to learn a distinct summarization pattern.
\end{abstract}

\maketitle

\section{Introduction}
\label{sec:introduction}

\begin{figure}[ht]
  \centering
  \includegraphics[width=0.6\textwidth]{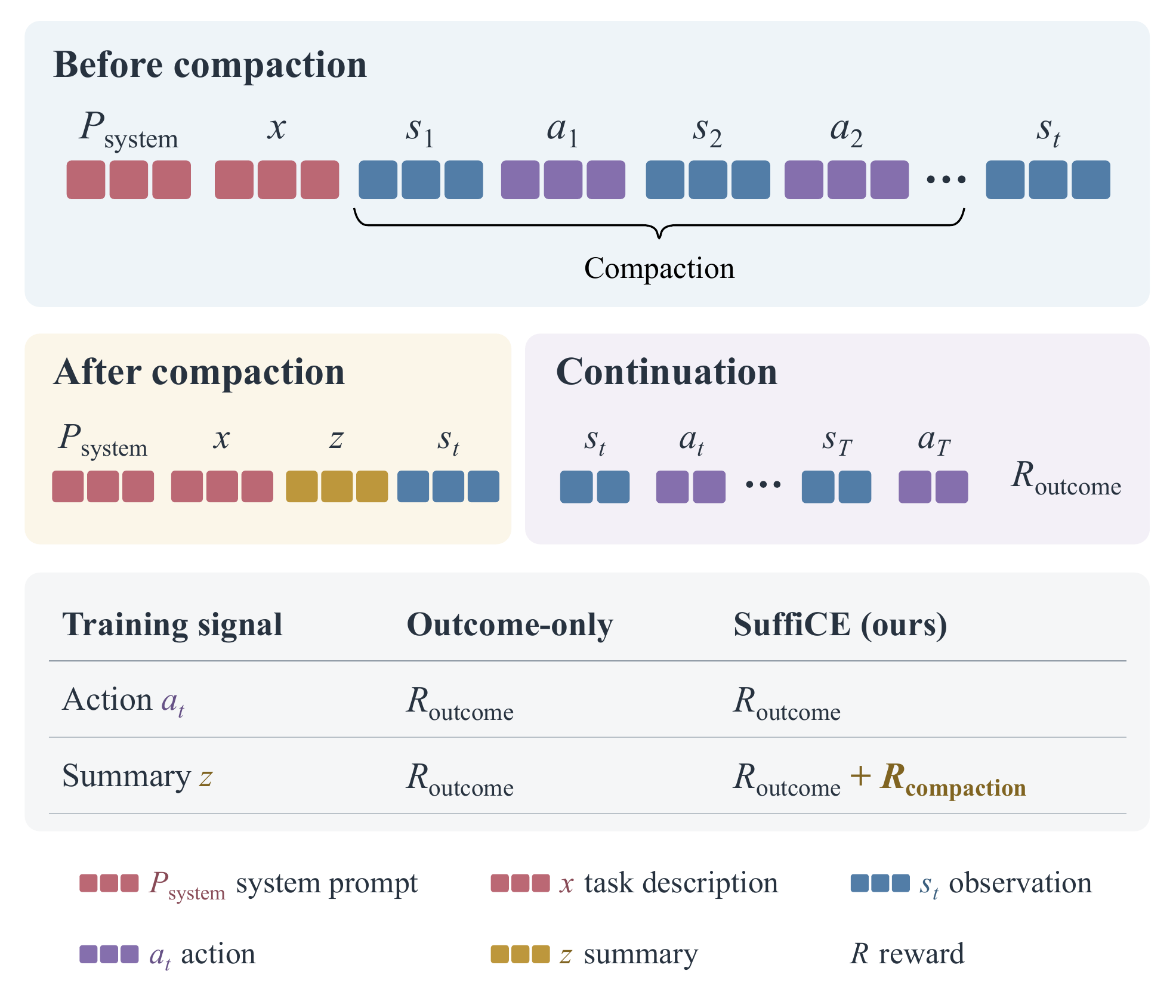}
  \caption{Credit assignment in compaction-integrated agentic RL. Context compaction replaces earlier interaction history with a summary $z$ so that the agent can continue interacting. Outcome-only training supervises both actions and summaries with this terminal reward alone, whereas SuffiCE adds a local compaction reward specifically for summary generation.}
  \label{fig:suffice_teaser}
\end{figure}

Agentic RL is an important post-training approach for LLMs. It trains LLM-based agents through interaction with real-world environments, using trial and error and task feedback to improve the agentic capabilities of the underlying models~\citep{zhang2025landscape}. This training enables agents to tackle increasingly complex reasoning, coding, and tool-use tasks~\citep{jin2025search,zeng2026glm,wei2025webagent}. As tasks become more complex, agents need more rounds of interaction to explore possible solutions and learn effective strategies. This paradigm scales by extending the interaction horizon, giving agents more rounds of interaction to explore and learn to solve increasingly complex long-horizon problems~\citep{xi2025agentgymrl}.

The context window, however, limits how far the interaction horizon can be extended. Each round adds actions and observations to the history, causing the context to grow approximately linearly with the number of interactions until the window is exhausted. Truncating a rollout interrupts the trajectory and can bias advantage estimates if an incomplete episode is treated as a completed one. Enlarging the context window substantially increases computational costs and does not guarantee that the model can effectively use the additional context~\citep{laban2026llms}. These constraints restrict exploration and limit the policy to learn from complete long-horizon interaction trajectories.

Context compaction is a widely used approach to overcoming this constraint. It replaces the accumulated interaction history with a concise representation, freeing space for subsequent interactions. Earlier approaches rely on external components, including predefined history-reduction procedures, parameterized memory representations, and dedicated summarization models~\citep{wang2025recursively,ge2023context,wu2025resum}. These components are not jointly optimized with the policy, so their ability to preserve useful information does not improve as the policy learns. The capabilities of the compaction module can therefore limit the task performance of the overall agentic system.

As base models have become more capable, recent work has incorporated compaction into the interaction process and jointly trained summary generation and decision making through agentic RL~\citep{lu2025scaling,li2026compactionrl}. As shown in Figure~\ref{fig:suffice_teaser}, these methods adapt advantage estimation to train across compaction boundaries, but still supervise summaries with sparse outcome rewards. Such rewards do not distinguish different sources of error. When a task fails, the outcome alone does not reveal whether the failure resulted from an incorrect decision or from information distorted in a summary. When initial summarization capabilities are weak, this ambiguity makes it difficult to learn effective summarization directly through RL. We refer to this difficulty as the summarization cold-start problem. SFT can provide an initial summarization capability, but requires additional supervision when adapted to new tasks~\citep{wu2025resum,wang2026self}. We therefore ask: \textit{how can we construct a reward that evaluates summary quality without additional supervision, so as to improve credit assignment and address the cold-start problem in compaction-integrated agentic RL?}

To provide such a reward, we propose \textbf{Suffi}x-\textbf{C}ompatibility \textbf{E}stimation (SuffiCE). The underlying idea is that a good summary should preserve the information needed for subsequent decisions, so replacing the full history with the summary should change the subsequent trajectory distribution as little as possible. SuffiCE combines deferred summarization with compatibility scoring. Deferred summarization delays the application of a generated summary while the agent continues to interact under the original context, collecting a local reference trajectory that does not depend on the summary. Compatibility scoring replays this reference window under the compacted context and compares its likelihoods under the original and compacted contexts to estimate the shift in the subsequent trajectory distribution. The resulting compatibility reward provides local feedback for summary generation, enabling more accurate credit assignment.

Across every benchmark and backbone, SuffiCE outperforms all compaction and non-compaction baselines, demonstrating that this local reward improves credit assignment and overcomes the summarization cold-start problem. With Qwen3-4B and Qwen2.5-7B-Instruct, SuffiCE reaches 87.6\% and 97.1\% success on ALFWorld, 84.4\% and 89.0\% score on WebShop, and 79.6\% and 85.1\% score on SciWorld, surpassing GRPO with compaction by 2.6 to 7.3 percentage points on these metrics. Further ablations on SciWorld validate our design choices, and embedding analysis of the generated summaries shows that compatibility rewards drive the policy to learn a distinct summarization pattern. Besides, SuffiCE increases per-step training time by only 5.7\% over the baseline.

\section{Related Work}
\label{sec:related_work}

\subsection{LLM-based Agents}
\label{sec:llm_based_agents}

An LLM-based agent uses a harness to drive an LLM, equipping it with modules such as tools and memory so that it can interact with the real world and solve practical problems~\citep{wang2024survey,xi2025rise}. Early agentic systems focus on system design, building components for prompting, planning, reflection, and memory~\citep{karpas2022mrkl,xu2023rewoo,yao2023react,wang2024executable,shinn2023reflexion,kim2025reflact,erdogan2025plan,wang2023voyager,park2023generative}. To improve agents more effectively, recent work increasingly turns to training the foundation model and to recursive self-improvement (RSI), in which agents iteratively optimize their own prompts, workflows, or code~\citep{schick2023toolformer,qin2024toolllm,patil2024gorilla,chen2023fireact,zeng2024agenttuning,chen2024agent,song2024agentbank,ou2025symbolic,khattab2024dspy,hu2025automated,zhang2026darwin}. Another line of work explores multi-agent systems to further improve performance~\citep{wu2024autogen,hong2024metagpt,qian2024chatdev,du2024improving,zhuge2024gptswarm,xue2025comas}. Agents are now widely applied in scenarios such as embodied tasks, office assistance, software engineering, and scientific research~\citep{shridhar2021alfworld,ahn2022can,yao2022webshop,zhou2024webarena,xie2024osworld,yang2024swe,wang2022scienceworld,lu2024ai,xu2025probing}. In this work, we post-train the model to address the limited context of agents on long-horizon tasks.

\subsection{Context Compaction}
\label{sec:context_compaction}

Limited computation prevents the context of agents from growing indefinitely~\citep{liu2024lost,hsieh2024ruler}. Context compaction therefore compresses the history with minimal information loss near the window limit, allowing agents to continue their tasks~\citep{li2025prompt,zhang2025survey}. Prior work explores rule-based rewriting systems and parametric representations of information~\citep{jiang2023llmlingua,jiang2024longllmlingua,pan2024llmlingua,lindenbauer2025complexity,wang2025recursively,mu2023learning,ge2023context}. As foundation models become more capable, LLM-generated summarization has also become a viable route~\citep{lee2024human,packer2023memgpt,hu2025hiagent,li2026self}. Meanwhile, many methods have been designed to improve the quality of compaction~\citep{yu2026memagent,kang2025acon,wu2025resum,ning2026agent,yi2026learning,li2026escaping}. A recent and promising direction lets the policy model generate its own summaries, so that decision making and summary generation can be jointly optimized through post-training~\citep{lu2025scaling,li2026compactionrl,wang2026self,sun2026scaling,hu2026ziprl,li2026mempo}. However, these methods remain immature, facing a difficult cold start and poor generalization. Our work follows this direction and improves the joint learning of decision making and summary generation through agentic RL.

\subsection{Agentic Reinforcement Learning}
\label{sec:agentic_reinforcement_learning}

RL for LLMs started from reinforcement learning from human feedback (RLHF), which trains LLMs to align with human preferences~\citep{schulman2017proximal,ziegler2019fine,ouyang2022training}. Many methods have since improved on this paradigm~\citep{rafailov2023direct,azar2024general,ethayarajh2024kto,ahmadian2024back,hu2025reinforce}. With the emergence of reinforcement learning with verifiable rewards (RLVR), RL has been widely used to improve the reasoning capabilities of LLMs~\citep{shao2024deepseekmath,guo2025deepseek}. A series of algorithms has followed to further improve RL training in this setting~\citep{liu2025understanding,lin2025cppo,yu2025dapo,zheng2025group,kazemnejad2024vineppo}. As LLM-based agents emerge, agentic RL, which optimizes policies with rewards from environment interaction, has become a new mainstream paradigm~\citep{zhang2025landscape}. A large body of work builds the infrastructure, frameworks, and environments for this pipeline~\citep{sheng2025hybridflow,rllm2025,luo2025agent,cheng2025agent,xi2025agentgym,xi2025agentgymrl}. The main challenges of agentic RL are exploration and credit assignment, which many methods aim to address~\citep{wang2025ragen,wang2026ragen,feng2025towards,zhang2026crps,ji2025tree,zhao2026branpo,zhou2024archer,feng2025group,peng2026hiper,he2026hierarchy,xue2026strata}. In this work, we further address credit assignment for summaries in compaction-integrated agentic RL by providing each summary with a local reward.

\section{Preliminaries}
\label{sec:preliminaries}

\subsection{Trajectories and Working Context}
\label{sec:trajectories}

LLM-based agents complete tasks through multi-step interaction with an environment. We denote a trajectory with $T$ task actions as:
\begin{equation}
    \tau = (x, s_1, a_1, s_2, \ldots, a_T, R^{\mathrm{outcome}}),
    \label{eq:trajectory}
\end{equation}
where $x$ is the global task description that stays in the context throughout the episode, $s_t$ is the environment observation before the $t$-th task action, $a_t$ is that action, and $R^{\mathrm{outcome}}$ is the terminal task return. Each action is a token sequence that may contain reasoning tokens and an environment command. Within a trajectory, the policy $\pi_\theta$ generates each action from the working context $H_t$, starting from $H_1 = [P_{\mathrm{system}}, x, s_1]$, where $P_{\mathrm{system}}$ is the system instruction. Without compaction, the agent acts and updates its context as:
\begin{equation}
    a_t \sim \pi_\theta(\cdot \mid H_t), \quad H_{t+1} = [H_t, a_t, s_{t+1}],
    \label{eq:context_update}
\end{equation}
where $[\cdot]$ denotes the concatenation of context segments. As actions and observations accumulate, the working context keeps growing and may eventually exceed the model's context window.

\subsection{Context Compaction}
\label{sec:compaction}

We only consider context compaction triggered at a fixed context length. Once the working context $H_t$ reaches $B_{\mathrm{trigger}}$ tokens, the system asks the policy to summarize the interaction history:
\begin{equation}
    z \sim \pi_\theta\left(\cdot \mid [H_t \setminus s_t, P_{\mathrm{compact}}]\right),
    \label{eq:summary_generation}
\end{equation}
where $z$ is the generated summary, $P_{\mathrm{compact}}$ is the compaction instruction, and $H_t \setminus s_t$ denotes $H_t$ without the current observation $s_t$, which is kept verbatim after compaction. The system then replaces the preceding interaction history as:
\begin{equation}
    H'_t = [P_{\mathrm{system}}, x, z, s_t].
    \label{eq:compressed_context}
\end{equation}
The agent generates $a_t$ from the compressed context $H'_t$ and continues to interact with the environment. We call the context $H_t$ before this compaction the raw context. It may itself contain summaries from earlier compactions, so compactions can be nested within an episode.

\subsection{GRPO-based Agentic Policy Optimization}
\label{sec:grpo}

We focus on GRPO-style methods~\citep{shao2024deepseekmath}. Given a task $x$, GRPO repeatedly samples $N$ trajectories to form a group $G$ and normalizes their terminal returns within the group to obtain the advantage of each trajectory:
\begin{equation}
    A_i^{\mathrm{outcome}} = \frac{R_i^{\mathrm{outcome}} - \operatorname{mean}(\{R_j^{\mathrm{outcome}}\}_{j=1}^{N})}{\operatorname{std}(\{R_j^{\mathrm{outcome}}\}_{j=1}^{N})}.
    \label{eq:return_advantage}
\end{equation}
Typically, $A_i^{\mathrm{outcome}}$ is shared by all tokens in the $i$-th trajectory. For a response $y$ generated from context $c$ and assigned advantage $A$, GRPO computes the clipped surrogate:
\begin{equation}
    \mathcal{L}(y, A; \theta) = \sum_{k=1}^{|y|} \min\left(\rho_k(\theta) A,\; \operatorname{clip}\left(\rho_k(\theta), 1 - \epsilon_{\mathrm{low}}, 1 + \epsilon_{\mathrm{high}}\right) A\right),
    \label{eq:clipped_surrogate}
\end{equation}
where $\rho_k(\theta) = \frac{\pi_\theta(y_k \mid c, y_{<k})}{\pi_{\theta_{\mathrm{old}}}(y_k \mid c, y_{<k})}$ is the importance ratio between the current policy $\pi_\theta$ and the sampling policy $\pi_{\theta_{\mathrm{old}}}$, and $\epsilon_{\mathrm{low}}$ and $\epsilon_{\mathrm{high}}$ are the clipping bounds~\citep{schulman2017proximal,yu2025dapo}. In agentic RL, we treat each user-assistant exchange as a turn and let $G_{i,j}$ denote the response in the $j$-th turn of the $i$-th trajectory. We first average the loss over all generated tokens within each trajectory and then average over the trajectories in the group, which gives the agentic GRPO objective:
\begin{equation}
    \mathcal{J}_{\mathrm{GRPO}}(\theta) = \mathbb{E}_{x \sim \mathcal{D},\, G \sim \pi_\theta(\cdot \mid x)}\left[\frac{1}{N}\sum_{i=1}^{N}\frac{1}{\sum_{j=1}^{|G_i|}|G_{i,j}|}\sum_{j=1}^{|G_i|}\mathcal{L}\left(G_{i,j}, A_i^{\mathrm{outcome}}; \theta\right)\right],
    \label{eq:grpo_objective}
\end{equation}
where $\mathcal{D}$ is the training task distribution, $|G_i|$ is the number of turns in the $i$-th trajectory, and $|G_{i,j}|$ is the number of tokens in $G_{i,j}$. Following common practice, we omit the KL penalty~\citep{yu2025dapo}. In this way, GRPO reinforces behaviors with high relative advantage within the group and suppresses those with low relative advantage.

\section{Method}
\label{sec:method}

\begin{figure}[ht]
    \centering
    \includegraphics[width=\textwidth]{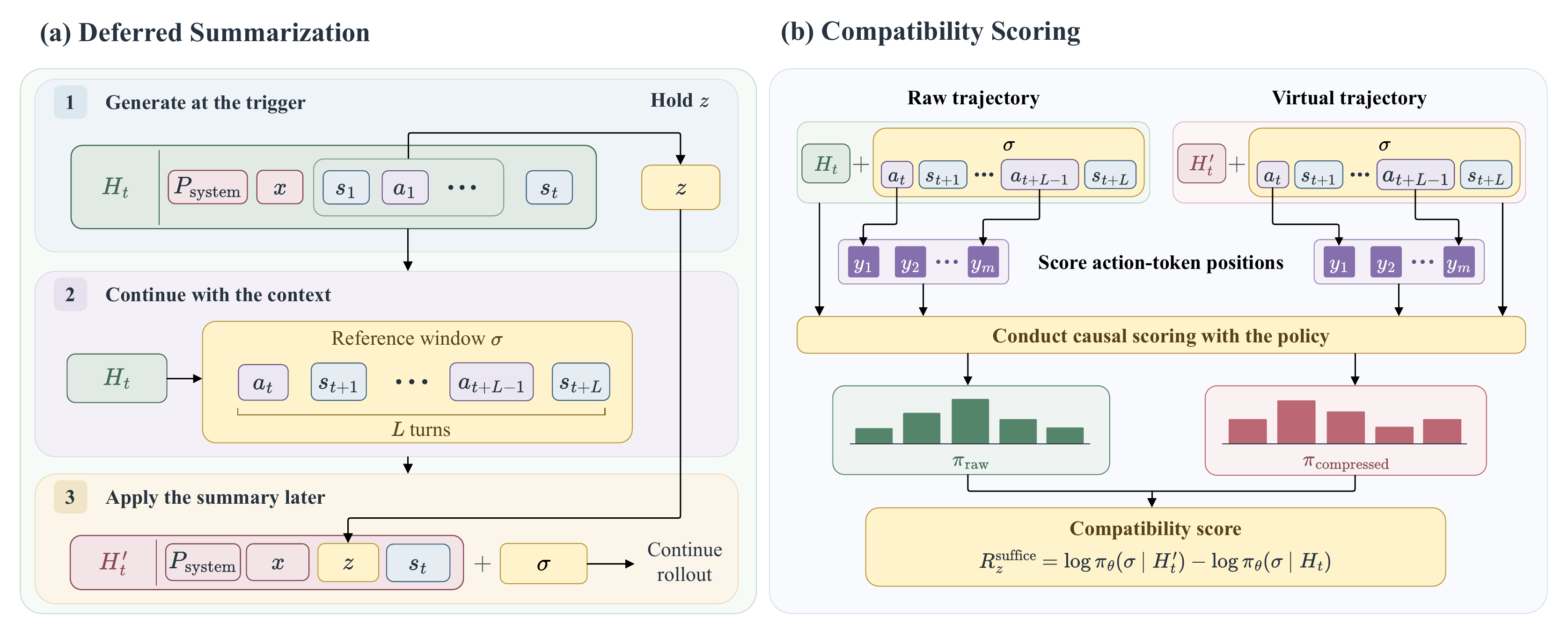}
    \caption{The components of SuffiCE. (a) Deferred summarization holds the summary $z$ while the agent acts under the raw context $H_t$ to collect a reference window $\sigma$, and then applies $z$ while retaining $\sigma$ verbatim. (b) Compatibility scoring evaluates $\sigma$ under both the raw trajectory and a virtual trajectory from the compressed context $H'_t$, using their log-likelihood difference $R_z^{\mathrm{suffice}}$ to measure the shift in the subsequent trajectory distribution.}
    \label{fig:suffice_method}
\end{figure}

SuffiCE builds on the intuition that a good summary should retain the information needed by subsequent decisions, so replacing the raw context $H_t$ with the compressed context $H'_t$ preserves the trajectory distribution. A natural measure of this difference is the KL divergence from the raw-context distribution to the compressed-context distribution:
\begin{equation}
    D_{\mathrm{KL}}\big(\pi_\theta(\cdot \mid H_t)\,\Vert\,\pi_\theta(\cdot \mid H'_t)\big) = \mathbb{E}_{\tau_{>t} \sim \pi_\theta(\cdot \mid H_t)}\left[\log\frac{\pi_\theta(\tau_{>t} \mid H_t)}{\pi_\theta(\tau_{>t} \mid H'_t)}\right],
    \label{eq:trajectory_kl}
\end{equation}
where $\tau_{>t} = (a_t, s_{t+1}, a_{t+1}, \ldots)$ is the subsequent trajectory from step $t$ onward, and $\pi_\theta(\cdot \mid H_t)$ and $\pi_\theta(\cdot \mid H'_t)$ are the subsequent trajectory distributions induced by the raw and compressed contexts.

Computing this KL divergence over the trajectory space is intractable, and repeatedly sampling from a given history requires rewinding the environment at high cost. We therefore adopt pair-wise estimation. We sample a subsequent trajectory $\tau_{>t}$ from $\pi_\theta(\cdot \mid H_t)$, replay it verbatim under $H'_t$ to compute $\pi_\theta(\tau_{>t} \mid H'_t)$, and use their log-likelihood ratio as an unbiased estimate. This unbiasedness is proved in Appendix~\ref{app:kl_unbiasedness}. As shown in Figure~\ref{fig:suffice_method}, SuffiCE uses deferred summarization to sample a local trajectory segment under the raw context, and compatibility scoring to replay it for a local compatibility reward in policy optimization.

\subsection{Deferred Summarization}
\label{sec:deferred}

In typical context compaction, a summary is applied immediately and subsequent turns are sampled from $\pi_\theta(\cdot \mid H'_t)$. In contrast, when $H_t$ reaches $B_{\mathrm{trigger}}$ at step $t$, we still generate $z$ via Equation~\ref{eq:summary_generation} but defer applying it while the agent executes $L$ more actions under $H_t$ to collect a reference window:
\begin{equation}
    \sigma = (a_t, s_{t+1}, a_{t+1}, \ldots, a_{t+L-1}, s_{t+L}).
    \label{eq:reference_window}
\end{equation}
Fewer than $L$ actions may be collected if the generation budget runs out early. Letting $\sigma_{<\ell}$ be the first $\ell-1$ turns in $\sigma$, each action is sampled as:
\begin{equation}
    a_{t+\ell-1} \sim \pi_\theta(\cdot \mid [H_t, \sigma_{<\ell}]).
    \label{eq:window_sampling}
\end{equation}
Since $z$ is not yet applied, $\sigma$ forms a genuine local trajectory sample from $\pi_\theta(\cdot \mid H_t)$. Once $\sigma$ is collected, we apply $z$ and rebuild the context by appending $\sigma$ verbatim after $H'_t$:
\begin{equation}
    H'_{t+L} = [H'_t, \sigma] = [P_{\mathrm{system}}, x, z, s_t, \sigma],
    \label{eq:deferred_rebuild}
\end{equation}
where $z$ replaces the history before $s_t$. Deferring compaction wastes no environment interactions within the window, and the agent continues interacting from the rebuilt working context $H'_{t+L}$.

\subsection{Compatibility Scoring}
\label{sec:compatibility}

Given $\sigma$, SuffiCE replays it verbatim after $H'_t$ as a scoring-only virtual trajectory $[H'_t, \sigma]$ and computes its log-probability within a single forward pass. Taking the negative log-likelihood ratio gives the compatibility reward:
\begin{equation}
    R_z^{\mathrm{suffice}} = \log\pi_\theta(\sigma \mid H'_t) - \log\pi_\theta(\sigma \mid H_t).
    \label{eq:compatibility_reward}
\end{equation}
In expectation, $R_z^{\mathrm{suffice}} = 0$ means the summary fully preserves the policy's behavioral distribution, whereas the magnitude of $R_z^{\mathrm{suffice}} < 0$ measures the behavioral shift caused by compaction. Because unknown environment transitions cancel when replaying the same trajectory, we only need to compute the normalized log-probability over all the output tokens, averaging within each action and then across the window:
\begin{equation}
    \log\pi_\theta(\sigma \mid c) = \frac{1}{L}\sum_{\ell=1}^{L}\frac{1}{|a_{t+\ell-1}|}\sum_{k=1}^{|a_{t+\ell-1}|}\log\pi_\theta\big(a_{t+\ell-1,k} \mid [c, \sigma_{<\ell}], a_{t+\ell-1,<k}\big),
    \label{eq:window_logprob}
\end{equation}
where $c \in \{H_t, H'_t\}$ is the context. When $c = H_t$, the log-probability terms are taken directly from the sampled trajectory. When $c = H'_t$, they are computed from the virtual trajectory $[H'_t, \sigma]$.

However, optimizing $R_z^{\mathrm{suffice}}$ alone allows the policy to copy the interaction history verbatim as a trivial optimum with $R_z^{\mathrm{suffice}} = 0$, which is an undesirable form of reward hacking. We therefore add a piecewise-linear length budget reward:
\begin{equation}
    R_z^{\mathrm{budget}} = -\operatorname{clip}\left(\frac{|z| - B_{\mathrm{threshold}}}{B_{\mathrm{boundary}} - B_{\mathrm{threshold}}}, 0, 1\right),
    \label{eq:budget_reward}
\end{equation}
where $|z|$ is the number of tokens in the summary, $B_{\mathrm{threshold}}$ is the length boundary up to which the reward remains $0$, and $B_{\mathrm{boundary}}$ is the length boundary at which the reward reaches $-1$. Between these two boundaries, the length budget reward decreases linearly. Together, the two rewards encourage the policy to retain the information most critical for subsequent decisions within the length budget.

\subsection{Policy Optimization}
\label{sec:policy_optimization}

SuffiCE follows the GRPO-style policy optimization from Section~\ref{sec:grpo}, except that agent trajectories may contain both action turns and summary turns. For action turns, we retain the shared outcome advantage $A_i^{\mathrm{outcome}}$. For each summary $z$, we additionally compute a compatibility advantage and a length budget advantage:
\begin{equation}
    \begin{aligned}
        A_z^{\mathrm{suffice}} &= \operatorname{clip}\left(\frac{R_z^{\mathrm{suffice}} - \operatorname{mean}(\{R_{z'}^{\mathrm{suffice}}\}_{z' \in \mathcal{Z}})}{\operatorname{std}(\{R_{z'}^{\mathrm{suffice}}\}_{z' \in \mathcal{Z}})},\, -\kappa, \kappa\right), \\
        A_z^{\mathrm{budget}} &= \frac{R_z^{\mathrm{budget}} - \operatorname{mean}(\{R_{z'}^{\mathrm{budget}}\}_{z' \in \mathcal{Z}})}{\operatorname{std}(\{R_{z'}^{\mathrm{budget}}\}_{z' \in \mathcal{Z}})},
    \end{aligned}
    \label{eq:summary_local_advantages}
\end{equation}
where $\mathcal{Z}$ is the set containing all summaries generated within $G$. Since $R_z^{\mathrm{suffice}}$ is unbounded, we clip $A_z^{\mathrm{suffice}}$ to $[-\kappa, \kappa]$ for numerical stability. Because $\mathcal{Z}$ spans different trajectories within $G$, $A_z^{\mathrm{suffice}}$ and $A_z^{\mathrm{budget}}$ still provide non-zero supervision signals even when all trajectories fail early in training. We then combine the outcome advantage with a weighted sum of these two local advantages to obtain the total advantage for summary $z$:
\begin{equation}
    A_z^{\mathrm{summary}} = A_i^{\mathrm{outcome}} + \lambda_{\mathrm{suffice}} A_z^{\mathrm{suffice}} + \lambda_{\mathrm{budget}} A_z^{\mathrm{budget}},
    \label{eq:summary_advantage}
\end{equation}
where $A_i^{\mathrm{outcome}}$ is the outcome advantage of the trajectory containing $z$, and $\lambda_{\mathrm{suffice}}$ and $\lambda_{\mathrm{budget}}$ are the weights on the local advantages, controlling the strength of the local rewards. Putting these together yields the overall SuffiCE objective:
\begin{equation}
    \mathcal{J}_{\mathrm{SuffiCE}}(\theta) = \mathbb{E}_{x \sim \mathcal{D},\, G \sim \pi_\theta(\cdot \mid x)}\left[\frac{1}{N}\sum_{i=1}^{N}\frac{1}{\sum_{j=1}^{|G_i|}|G_{i,j}|}\sum_{j=1}^{|G_i|}\mathcal{L}\left(G_{i,j}, A_{i,j}; \theta\right)\right],
    \label{eq:suffice_objective}
\end{equation}
where $A_{i,j}$ denotes the advantage of the $j$-th turn in the $i$-th trajectory, equal to $A_i^{\mathrm{outcome}}$ for action turns and $A_z^{\mathrm{summary}}$ for summary turns.

\section{Experiments}
\label{sec:experiments}

\begin{table}[t]
    \caption{Performance on ALFWorld and WebShop. ALFWorld reports subtask and overall success rates. WebShop reports overall task score and success rate. Values are mean $\pm$ standard deviation over three seeds on the held-out test split. The \hlfirst{best} and \hlsecond{second-best} results are highlighted.}
    \label{tab:alfworld_and_webshop_results}
    \renewcommand{\arraystretch}{1.0}
    \centering \footnotesize \setlength{\tabcolsep}{1.5pt}
    \begin{tabular}{lccccccc|cc}
        \toprule
        \multirow{2}{*}{\textbf{Method}} & \multicolumn{7}{c|}{\textbf{ALFWorld}} & \multicolumn{2}{c}{\textbf{WebShop}} \\
        & Pick & Look & Clean & Heat & Cool & Pick2 & All & Score & Succ. \\
        \midrule
        \multicolumn{10}{l}{\textit{Qwen3-4B}} \\
        \quad Untrained w/o Comp. & $77.1_{\pm 2.3}$ & $35.9_{\pm 15.8}$ & $16.0_{\pm 4.6}$ & $4.2_{\pm 2.9}$ & $0.0_{\pm 0.0}$ & $25.0_{\pm 9.0}$ & $30.5_{\pm 2.4}$ & $12.0_{\pm 1.4}$ & $6.7_{\pm 1.1}$ \\
        \quad Untrained w/ Comp. & $87.6_{\pm 3.6}$ & $38.5_{\pm 6.3}$ & $32.1_{\pm 1.7}$ & $4.2_{\pm 2.9}$ & $5.3_{\pm 5.0}$ & $58.3_{\pm 3.4}$ & $43.1_{\pm 2.9}$ & $17.9_{\pm 0.9}$ & $7.4_{\pm 1.0}$ \\
        \quad PPO w/o Comp. & \hlfirst{$100.0_{\pm 0.0}$} & \hlfirst{$94.9_{\pm 3.6}$} & \hlsecond{$84.0_{\pm 4.6}$} & \hlsecond{$75.0_{\pm 0.0}$} & $65.3_{\pm 6.8}$ & $58.3_{\pm 3.4}$ & $80.2_{\pm 0.9}$ & \hlsecond{$83.3_{\pm 0.5}$} & $70.3_{\pm 0.4}$ \\
        \quad PPO w/ Comp. & \hlsecond{$100.0_{\pm 0.0}$} & $76.9_{\pm 6.3}$ & $64.2_{\pm 1.7}$ & $72.9_{\pm 2.9}$ & \hlsecond{$70.7_{\pm 6.8}$} & $80.6_{\pm 5.2}$ & $79.3_{\pm 0.6}$ & $78.4_{\pm 0.4}$ & $66.1_{\pm 0.4}$ \\
        \quad GRPO w/o Comp. & $98.1_{\pm 2.7}$ & $76.9_{\pm 6.3}$ & $60.5_{\pm 1.7}$ & $75.0_{\pm 5.1}$ & $69.3_{\pm 5.0}$ & $80.6_{\pm 3.9}$ & $78.1_{\pm 0.9}$ & $82.9_{\pm 0.7}$ & \hlsecond{$71.3_{\pm 0.6}$} \\
        \quad GRPO w/ Comp. & $100.0_{\pm 0.0}$ & \hlsecond{$82.1_{\pm 7.3}$} & $66.7_{\pm 6.0}$ & $68.8_{\pm 10.2}$ & $68.0_{\pm 3.3}$ & \hlsecond{$90.3_{\pm 8.6}$} & \hlsecond{$81.0_{\pm 0.9}$} & $77.1_{\pm 0.7}$ & $63.1_{\pm 0.4}$ \\
        \quad SuffiCE (Ours) & $96.2_{\pm 1.3}$ & $74.4_{\pm 3.6}$ & \hlfirst{$91.4_{\pm 1.7}$} & \hlfirst{$75.0_{\pm 0.0}$} & \hlfirst{$84.0_{\pm 6.5}$} & \hlfirst{$90.3_{\pm 3.9}$} & \hlfirst{$87.6_{\pm 2.0}$} & \hlfirst{$84.4_{\pm 0.2}$} & \hlfirst{$74.0_{\pm 0.2}$} \\
        \midrule
        \multicolumn{10}{l}{\textit{Qwen2.5-7B-Instruct}} \\
        \quad Untrained w/o Comp. & $25.7_{\pm 4.7}$ & $43.6_{\pm 3.6}$ & $11.1_{\pm 0.0}$ & $0.0_{\pm 0.0}$ & $5.3_{\pm 1.9}$ & $25.0_{\pm 3.4}$ & $17.9_{\pm 2.0}$ & $36.4_{\pm 0.9}$ & $23.6_{\pm 0.7}$ \\
        \quad Untrained w/ Comp. & $51.4_{\pm 2.3}$ & $59.0_{\pm 3.6}$ & $25.9_{\pm 3.0}$ & $25.0_{\pm 0.0}$ & $32.0_{\pm 3.3}$ & $36.1_{\pm 7.1}$ & $38.1_{\pm 0.9}$ & $39.2_{\pm 0.6}$ & $25.0_{\pm 0.7}$ \\
        \quad PPO w/o Comp. & $97.1_{\pm 0.0}$ & \hlfirst{$100.0_{\pm 0.0}$} & $92.6_{\pm 0.0}$ & $81.2_{\pm 0.0}$ & $84.0_{\pm 0.0}$ & $83.3_{\pm 0.0}$ & $90.0_{\pm 0.0}$ & $83.0_{\pm 0.1}$ & $72.3_{\pm 0.1}$ \\
        \quad PPO w/ Comp. & \hlsecond{$100.0_{\pm 0.0}$} & \hlsecond{$100.0_{\pm 0.0}$} & \hlsecond{$100.0_{\pm 0.0}$} & \hlsecond{$87.5_{\pm 0.0}$} & \hlsecond{$92.0_{\pm 3.3}$} & $91.7_{\pm 0.0}$ & \hlsecond{$95.7_{\pm 0.6}$} & $84.8_{\pm 0.4}$ & \hlsecond{$78.8_{\pm 0.3}$} \\
        \quad GRPO w/o Comp. & $97.1_{\pm 0.0}$ & $100.0_{\pm 0.0}$ & $96.3_{\pm 0.0}$ & $79.2_{\pm 2.9}$ & $77.3_{\pm 1.9}$ & $91.7_{\pm 0.0}$ & $90.7_{\pm 0.6}$ & $84.0_{\pm 0.0}$ & $72.7_{\pm 0.1}$ \\
        \quad GRPO w/ Comp. & $100.0_{\pm 0.0}$ & $94.9_{\pm 3.6}$ & $96.3_{\pm 0.0}$ & $87.5_{\pm 5.1}$ & $88.0_{\pm 0.0}$ & \hlsecond{$94.4_{\pm 3.9}$} & $94.3_{\pm 0.6}$ & \hlsecond{$86.4_{\pm 0.5}$} & $76.8_{\pm 0.3}$ \\
        \quad SuffiCE (Ours) & \hlfirst{$100.0_{\pm 0.0}$} & $94.9_{\pm 3.6}$ & \hlfirst{$100.0_{\pm 0.0}$} & \hlfirst{$95.8_{\pm 2.9}$} & \hlfirst{$94.7_{\pm 5.0}$} & \hlfirst{$94.4_{\pm 2.0}$} & \hlfirst{$97.1_{\pm 0.6}$} & \hlfirst{$89.0_{\pm 0.5}$} & \hlfirst{$79.3_{\pm 0.6}$} \\
        \bottomrule
    \end{tabular}
\end{table}

\begin{table}[t]
    \caption{Performance on SciWorld. We report task scores and subtask breakdowns. Values are mean $\pm$ standard deviation over three seeds. The \hlfirst{best} and \hlsecond{second-best} results are highlighted.}
    \label{tab:sciworld_results}
    \renewcommand{\arraystretch}{1.0}
    \centering \footnotesize \setlength{\tabcolsep}{3pt}
    \begin{tabular}{lcccccc}
        \toprule
        \multirow{2}{*}{\textbf{Method}} & \multicolumn{6}{c}{\textbf{SciWorld}} \\
        & Measure & Test-Cond & Find & Chem-Mix & Lifespan & All \\
        \midrule
        \multicolumn{7}{l}{\textit{Qwen3-4B}} \\
        \quad Untrained w/o Comp. & $10.4_{\pm 2.2}$ & $3.2_{\pm 0.4}$ & $27.1_{\pm 1.5}$ & $11.7_{\pm 0.2}$ & $13.9_{\pm 2.8}$ & $11.5_{\pm 0.7}$ \\
        \quad Untrained w/ Comp. & $15.6_{\pm 5.0}$ & $5.6_{\pm 0.9}$ & $29.4_{\pm 0.5}$ & $13.6_{\pm 0.0}$ & $37.8_{\pm 6.7}$ & $16.2_{\pm 1.1}$ \\
        \quad PPO w/o Comp. & $60.6_{\pm 1.9}$ & $70.4_{\pm 0.4}$ & $67.8_{\pm 2.1}$ & $14.3_{\pm 1.9}$ & \hlfirst{$88.9_{\pm 1.6}$} & $68.2_{\pm 0.6}$ \\
        \quad PPO w/ Comp. & \hlsecond{$62.2_{\pm 1.8}$} & \hlsecond{$76.8_{\pm 1.2}$} & $82.3_{\pm 4.7}$ & \hlfirst{$39.6_{\pm 8.6}$} & \hlsecond{$88.9_{\pm 5.7}$} & \hlsecond{$75.8_{\pm 1.3}$} \\
        \quad GRPO w/o Comp. & $58.0_{\pm 0.4}$ & $76.4_{\pm 1.0}$ & $48.8_{\pm 2.7}$ & $14.9_{\pm 0.9}$ & $85.6_{\pm 8.3}$ & $65.5_{\pm 1.0}$ \\
        \quad GRPO w/ Comp. & $57.1_{\pm 2.8}$ & $73.2_{\pm 1.0}$ & \hlsecond{$83.4_{\pm 2.0}$} & $27.4_{\pm 5.7}$ & $84.4_{\pm 5.7}$ & $72.8_{\pm 1.1}$ \\
        \quad SuffiCE (Ours) & \hlfirst{$62.9_{\pm 2.0}$} & \hlfirst{$80.4_{\pm 0.7}$} & \hlfirst{$91.3_{\pm 1.2}$} & \hlsecond{$32.9_{\pm 8.5}$} & $88.3_{\pm 2.4}$ & \hlfirst{$79.6_{\pm 0.3}$} \\
        \midrule
        \multicolumn{7}{l}{\textit{Qwen2.5-7B-Instruct}} \\
        \quad Untrained w/o Comp. & $27.8_{\pm 3.6}$ & $22.6_{\pm 1.0}$ & $16.7_{\pm 0.8}$ & $16.3_{\pm 6.8}$ & $27.2_{\pm 5.5}$ & $22.2_{\pm 0.5}$ \\
        \quad Untrained w/ Comp. & $27.7_{\pm 1.2}$ & $28.4_{\pm 1.1}$ & $24.6_{\pm 1.6}$ & $10.3_{\pm 0.9}$ & $30.0_{\pm 3.6}$ & $27.0_{\pm 0.5}$ \\
        \quad PPO w/o Comp. & $55.3_{\pm 0.5}$ & $84.0_{\pm 0.0}$ & $79.6_{\pm 1.0}$ & $18.6_{\pm 0.0}$ & \hlfirst{$100.0_{\pm 0.0}$} & $77.8_{\pm 0.3}$ \\
        \quad PPO w/ Comp. & $64.7_{\pm 2.3}$ & $78.3_{\pm 1.0}$ & \hlfirst{$90.3_{\pm 1.4}$} & \hlfirst{$72.3_{\pm 13.2}$} & \hlsecond{$95.6_{\pm 1.6}$} & $80.4_{\pm 1.5}$ \\
        \quad GRPO w/o Comp. & \hlsecond{$70.4_{\pm 0.0}$} & $85.7_{\pm 0.1}$ & $55.5_{\pm 0.5}$ & $30.6_{\pm 0.0}$ & $95.6_{\pm 1.6}$ & $74.9_{\pm 0.0}$ \\
        \quad GRPO w/ Comp. & $63.2_{\pm 2.1}$ & \hlfirst{$87.5_{\pm 0.2}$} & $85.9_{\pm 1.6}$ & $35.3_{\pm 6.6}$ & $84.4_{\pm 1.6}$ & \hlsecond{$81.5_{\pm 0.1}$} \\
        \quad SuffiCE (Ours) & \hlfirst{$81.6_{\pm 1.9}$} & \hlsecond{$86.3_{\pm 0.9}$} & \hlsecond{$89.9_{\pm 0.6}$} & \hlsecond{$43.9_{\pm 10.0}$} & $84.4_{\pm 3.1}$ & \hlfirst{$85.1_{\pm 0.5}$} \\
        \bottomrule
    \end{tabular}
\end{table}

\subsection{Experimental Setup}
\label{sec:experimental_setup}

\noindent\textbf{Benchmarks.} We evaluate our method on three agentic benchmarks, including ALFWorld, WebShop, and SciWorld accessed through AgentGym~\citep{shridhar2021alfworld,yao2022webshop,wang2022scienceworld,xi2025agentgym}. We report the success rate and subtask breakdown on ALFWorld, the task score and success rate on WebShop, and the task score and subtask breakdown on SciWorld.

\noindent\textbf{Baselines.} We compare SuffiCE against two types of baselines. The first type consists of methods without compaction, including policies trained with PPO and GRPO, termed PPO w/o Comp. and GRPO w/o Comp.~\citep{schulman2017proximal,shao2024deepseekmath}. They truncate trajectories once they exceed the context budget. The second type comprises methods with compaction, triggering summarization at $B_{\mathrm{trigger}}$, including reproduced adaptations of CompactionRL and SUPO under PPO and GRPO, termed PPO w/ Comp. and GRPO w/ Comp.~\citep{li2026compactionrl,lu2025scaling}.

\noindent\textbf{Training Details.} Our codebase builds on verl and rLLM~\citep{sheng2025hybridflow,rllm2025}. We adopt Qwen3-4B and Qwen2.5-7B-Instruct as the backbone models~\citep{yang2025qwen3,yang2024qwen}. All methods train for 150 steps with a batch size of 16, where GRPO-style methods sample $N=8$ trajectories per task. SuffiCE sets the reference window length to $L=3$, the length budget thresholds to $B_{\mathrm{threshold}}=512$ and $B_{\mathrm{boundary}}=1024$, the local advantage weights to $\lambda_{\mathrm{suffice}}=\lambda_{\mathrm{budget}}=1.0$, and the advantage clipping threshold to $\kappa=3.0$. Further details are listed in Appendices~\ref{app:budget_settings} and~\ref{app:experimental_settings}. At evaluation, SuffiCE retains deferred summarization while disabling compatibility scoring, incurring no extra overhead. Each experiment runs on a single node with 8 NVIDIA H100 GPUs.

\subsection{Main Results}
\label{sec:main_results}

Tables~\ref{tab:alfworld_and_webshop_results} and~\ref{tab:sciworld_results} present the main results across the three benchmarks. RL training substantially improves over the untrained policies on every benchmark, and methods with compaction generally outperform their non-compacting counterparts. For example, with Qwen2.5-7B-Instruct, compaction lifts untrained ALFWorld success from 17.9\% to 38.1\% and improves GRPO from 90.7\% to 94.3\%. This gain arises because compaction prevents long trajectories from being truncated early, allowing the policy to explore thoroughly within the step budget and improving task success and sample efficiency in the early stage of training.

However, enabling compaction in PPO and GRPO yields unstable gains. With Qwen3-4B on WebShop, compaction lowers the task score from 83.3\% to 78.4\% for PPO and from 82.9\% to 77.1\% for GRPO. This indicates that without SFT, summary generation suffers from a cold-start problem. In contrast, SuffiCE surpasses all baselines across benchmarks and backbones with the best overall results, reaching 87.6\% and 97.1\% success on ALFWorld, 84.4\% and 89.0\% score on WebShop, and 79.6\% and 85.1\% score on SciWorld, showing that compatibility rewards overcome the cold-start problem and effectively promote the joint learning of decision making and summary generation.

\subsection{Ablation Studies}
\label{sec:ablation_studies}

\begin{table}[ht]
    \caption{Results of the summary strategy ablation on SciWorld with Qwen2.5-7B-Instruct. We report the average task score and number of steps, compactions, and tokens replacing the history. No comp. and Self Sum. refer to GRPO w/o Comp.\ and GRPO w/ Comp., Keep last and Keep recent retain the latest one and three turns verbatim, and Frozen Sum. uses the frozen initial model to summarize.}
    \label{tab:strategy_ablations}
    \renewcommand{\arraystretch}{1.0}
    \centering \footnotesize \setlength{\tabcolsep}{6pt}
    \begin{tabular}{lcccc}
        \toprule
        \textbf{Method} & \textbf{\%Score} & \textbf{\#Step} & \textbf{\#Comp.} & \textbf{\#Token} \\
        \midrule
        \multicolumn{5}{l}{\textit{Rule-based}} \\
        \quad No comp. & $74.9_{\pm 0.0}$ & $10.2$ & $0.0$ & $0.0$ \\
        \quad Keep last & $81.2_{\pm 0.2}$ & $19.4$ & $0.5$ & $157.8$ \\
        \quad Keep recent & $80.7_{\pm 0.1}$ & $24.3$ & $0.8$ & $449.4$ \\
        \midrule
        \multicolumn{5}{l}{\textit{Model-based}} \\
        \quad Frozen Sum. & $75.6_{\pm 0.4}$ & $25.9$ & $0.8$ & $60.9$ \\
        \quad Self Sum. & \hlsecond{$81.5_{\pm 0.1}$} & $21.7$ & $0.8$ & $60.1$ \\
        \quad SuffiCE (Ours) & \hlfirst{$85.1_{\pm 0.5}$} & $22.2$ & $0.5$ & $157.1$ \\
        \bottomrule
    \end{tabular}
\end{table}

Table~\ref{tab:strategy_ablations} presents the summary strategy ablation results on SciWorld with Qwen2.5-7B-Instruct, comparing rule-based retention with model-based summaries. Rule-based summaries achieve performance comparable to the compaction baseline, showing that merely introducing compaction into GRPO cannot overcome the cold-start problem. The frozen summarizer lacks task-specific knowledge and cannot co-evolve with the policy, thus failing to improve over the non-compaction baseline. Hence, SuffiCE proves essential for jointly learning decision making and summary generation. Another ablation study on the rationality of method design is provided in Appendix~\ref{app:method_design_ablation}.

\subsection{Behavior Visualization}
\label{sec:behavior_visualization}

\begin{figure}[ht]
  \centering
  \includegraphics[width=0.6\textwidth]{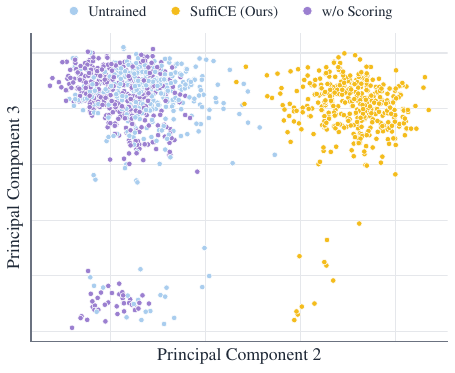}
  \caption{PCA projection of summary embeddings on SciWorld with Qwen2.5-7B-Instruct, comparing the untrained policy, SuffiCE, and its non-scoring variant. The second and third principal components are plotted, as the first primarily separates task categories.}
  \label{fig:summary_embedding}
\end{figure}

To directly illustrate how SuffiCE improves summary generation, we randomly sample summaries from the untrained policy, SuffiCE, and its non-scoring variant, and encode them into sentence embeddings using all-MiniLM-L6-v2~\citep{wang2020minilm}. Figure~\ref{fig:summary_embedding} shows their distribution in the principal component space~\citep{hotelling1933analysis}. While the non-scoring variant remains mixed with the untrained policy, SuffiCE forms a clearly separated cluster. This indicates that compatibility rewards drive the policy to learn a distinct summarization pattern. Further analysis on compaction behavior and quality is provided in Appendices~\ref{app:compaction_behavior} and~\ref{app:compaction_quality}.

\subsection{Computational Efficiency}
\label{sec:computational_efficiency}

\begin{figure}[ht]
  \centering
  \includegraphics[width=0.6\textwidth]{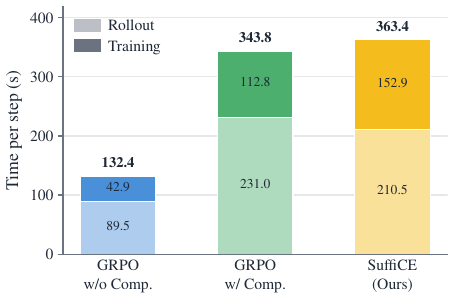}
  \caption{Per-step wall-clock time on SciWorld with Qwen2.5-7B-Instruct for SuffiCE and GRPO baselines. Each bar splits rollout and training time in lighter and darker shades.}
  \label{fig:training_time_step}
\end{figure}

Figure~\ref{fig:training_time_step} compares the per-step wall-clock time of SuffiCE and the GRPO baselines, broken down into the rollout and training stages. Compared with the non-compaction baseline, enabling compaction substantially increases the step time, as compaction allows longer trajectories whose computational cost grows with trajectory length. Relative to the compaction baseline, SuffiCE incurs additional yet acceptable overhead in the training stage to compute compatibility rewards, but takes less time in rollout owing to shorter trajectories. Consequently, the overall per-step time remains comparable with only a 5.7\% overhead. A detailed breakdown is provided in Appendix~\ref{app:time_consumption_breakdown}.

\section{Conclusion}
\label{sec:conclusion}

Although compaction-integrated agentic RL overcomes context limits, policies trained solely with sparse outcome rewards face a cold-start problem. To address this, we propose SuffiCE, built on the intuition that a good summary preserves the subsequent trajectory distribution. Deferred summarization collects a reference window under the raw context, and compatibility scoring compares its log-probabilities under the raw and compressed contexts to construct a compatibility reward, providing native local supervision for summary generation. Experiments on various benchmarks and backbones show that SuffiCE consistently outperforms all non-compaction and compaction baselines. Moreover, ablations verify the soundness and efficacy of our design, efficiency analysis shows overall per-step training time comparable to the compaction baseline, and behavior visualization further confirms that the policy develops an effective summarization pattern. Overall, SuffiCE provides an effective solution to the cold-start problem in compaction-integrated agentic RL. Future work can extend it to longer-horizon domains and explore more sophisticated compaction mechanisms.

\section*{Limitations}
\label{sec:limitations}

Although SuffiCE is concise and self-contained in design, it still has two potential limitations. First, although the pair-wise log-likelihood ratio in compatibility scoring provides an unbiased estimate of the KL divergence between subsequent trajectory distributions, using a single reference trajectory per compaction can incur high variance due to the limited sample size. This design reflects a practical trade-off against computational and interaction costs, and future work could explore variance-reduction methods such as control-variate KL estimators or exact per-token KL divergence over the full vocabulary distribution. Second, deferred summarization requires triggering compaction early to reserve space for the reference window under the raw context and retains this window verbatim upon reconstruction. This extra overhead lowers the agent's effective context window ceiling, motivating future work on adaptive window lengths or more context-efficient deferral mechanisms.

\section*{Ethical Considerations}
\label{sec:ethical_considerations}

All code, models, and environments are publicly available, benign open-source research resources, and our experiments involve no human subjects or private data, leaving only two minor ethical considerations. First, context compaction requires the model to summarize its own interaction history. This lossy process can exacerbate hallucination, as fabricated details or omitted constraints in a summary may compound over subsequent turns, placing higher demands on factual grounding and safety alignment. Second, SuffiCE's compatibility reward only encourages summaries to preserve the subsequent trajectory distribution under the raw context without eliminating inherent biases or unsafe tendencies in the underlying model. Therefore, real-world deployment should be paired with stronger safety alignment, non-compressible safety constraints, and necessary human oversight.

\bibliography{references}

\appendix

\section{Unbiasedness of KL Estimation}
\label{app:kl_unbiasedness}

Consider a partially observable environment with transition function $\mathcal{T}(u_{t+1}, s_{t+1} \mid u_t, a_t)$, where $u_t \in \mathcal{U}$ is the state before $a_t$~\citep{kaelbling1998planning}. Compaction at step $t$ replaces the working context $H_t$ with $H'_t$ while leaving $u_t$ unchanged. Let $\tau_{t:T} = (a_t, s_{t+1}, \ldots, a_{T-1}, s_T)$ denote the segment before $a_T$, which is empty if $T = t$. When $T \ge t + L$, we write $\tau_{>t} = (\sigma, \tau_{>t+L})$. We define the marginal observation distribution as:
\begin{equation}
    P_\epsilon(s_{t+1:T} \mid u_t, a_{t:T-1}) \triangleq \sum_{u_{t+1:T}} \prod_{i=t}^{T-1} \mathcal{T}(u_{i+1}, s_{i+1} \mid u_i, a_i),
    \label{eq:app_marginal_obs}
\end{equation}
which is independent of the context $c \in \{H_t, H'_t\}$. Since the policy observes only $[c, \tau_{t:i}]$ rather than $u_{t+1:T}$, marginalizing out $u_{t+1:T}$ gives the subsequent trajectory distribution:
\begin{equation}
    P_\theta(\tau_{>t} \mid u_t, c) = P_\epsilon(s_{t+1:T} \mid u_t, a_{t:T-1}) \cdot \prod_{i=t}^{T} \pi_\theta\big(a_i \mid [c, \tau_{t:i}]\big),
    \label{eq:app_full_factorization}
\end{equation}
and its $L$-step counterpart over $\sigma$:
\begin{equation}
    P_\theta^{(L)}(\sigma \mid u_t, c) = P_\epsilon(s_{t+1:t+L} \mid u_t, a_{t:t+L-1}) \cdot \prod_{\ell=1}^{L} \pi_\theta\big(a_{t+\ell-1} \mid [c, \sigma_{<\ell}]\big).
    \label{eq:app_window_factorization}
\end{equation}
Then define the turn-level log-likelihood ratio as:
\begin{equation}
    \Delta_i(\tau_{>t}) \triangleq \sum_{k=1}^{|a_i|} \log\frac{\pi_\theta\big(a_{i,k} \mid [H_t, \tau_{t:i}], a_{i,<k}\big)}{\pi_\theta\big(a_{i,k} \mid [H'_t, \tau_{t:i}], a_{i,<k}\big)}.
    \label{eq:app_turn_llr}
\end{equation}
Finally, define the full-trajectory and local-window pair-wise estimators as:
\begin{equation}
    \begin{aligned}
        \widehat{D}_{\mathrm{KL}}(\tau_{>t}; H_t, H'_t) &\triangleq \sum_{i=t}^{T} \Delta_i(\tau_{>t}), \\
        \widehat{D}_{\mathrm{KL}}^{(L)}(\sigma; H_t, H'_t) &\triangleq \sum_{\ell=1}^{L} \Delta_{t+\ell-1}(\sigma).
    \end{aligned}
    \label{eq:app_estimators}
\end{equation}

\begin{theorem}[Unbiasedness of Full-trajectory Estimation]
\label{thm:ideal_unbiased}
Suppose $\tau_{>t}$ is rolled out from $(u_t, H_t)$ to horizon $T \ge t$ with $z$ withheld and without context truncation or nested compactions. Then:
\begin{equation}
    \mathbb{E}_{\tau_{>t} \sim P_\theta(\cdot \mid u_t, H_t)}\big[\widehat{D}_{\mathrm{KL}}(\tau_{>t}; H_t, H'_t)\big] = D_{\mathrm{KL}}\big(P_\theta(\cdot \mid u_t, H_t)\,\Vert\,P_\theta(\cdot \mid u_t, H'_t)\big).
    \label{eq:app_ideal_unbiased}
\end{equation}
\end{theorem}

\begin{proof}
Since $z$ is withheld, the sampled trajectory follows $P_\theta(\cdot \mid u_t, H_t)$. Substituting Equation~\ref{eq:app_full_factorization} into the log-likelihood ratio cancels the marginal observation distribution $P_\epsilon(s_{t+1:T} \mid u_t, a_{t:T-1})$, and expanding $\pi_\theta(a_i \mid [c, \tau_{t:i}])$ gives:
\begin{equation}
    \log\frac{P_\theta(\tau_{>t} \mid u_t, H_t)}{P_\theta(\tau_{>t} \mid u_t, H'_t)} = \sum_{i=t}^{T} \log\frac{\pi_\theta\big(a_i \mid [H_t, \tau_{t:i}]\big)}{\pi_\theta\big(a_i \mid [H'_t, \tau_{t:i}]\big)} = \widehat{D}_{\mathrm{KL}}(\tau_{>t}; H_t, H'_t).
    \label{eq:app_full_llr}
\end{equation}
Taking the expectation over $\tau_{>t} \sim P_\theta(\cdot \mid u_t, H_t)$ yields the desired equality.
\end{proof}

\begin{theorem}[Unbiasedness of Local-window Estimation]
\label{thm:local_unbiased}
Suppose $\tau_{>t}$ is rolled out from $(u_t, H_t)$ to horizon $T \ge t + L$ with $z$ withheld and without context truncation or nested compactions. Besides, assume the history primarily governs immediate actions within $\sigma$, i.e., $\pi_\theta(\cdot \mid [H_t, \sigma, \tau_{t+L:i}]) = \pi_\theta(\cdot \mid [H'_t, \sigma, \tau_{t+L:i}])$ holds for all $t + L \leq i \leq T$~\citep{kara2022near,golowich2023planning}. Then:
\begin{equation}
    \begin{aligned}
        \mathbb{E}_{\sigma \sim P_\theta^{(L)}(\cdot \mid u_t, H_t)}\big[\widehat{D}_{\mathrm{KL}}^{(L)}(\sigma; H_t, H'_t)\big] &= D_{\mathrm{KL}}\big(P_\theta^{(L)}(\cdot \mid u_t, H_t)\,\Vert\,P_\theta^{(L)}(\cdot \mid u_t, H'_t)\big) \\
        &= D_{\mathrm{KL}}\big(P_\theta(\cdot \mid u_t, H_t)\,\Vert\,P_\theta(\cdot \mid u_t, H'_t)\big).
    \end{aligned}
    \label{eq:app_local_unbiased}
\end{equation}
\end{theorem}

\begin{proof}
Since $z$ is withheld, we have $\sigma \sim P_\theta^{(L)}(\cdot \mid u_t, H_t)$. Canceling the marginal observation distribution $P_\epsilon(s_{t+1:t+L} \mid u_t, a_{t:t+L-1})$ in Equation~\ref{eq:app_window_factorization} yields $\log\frac{P_\theta^{(L)}(\sigma \mid u_t, H_t)}{P_\theta^{(L)}(\sigma \mid u_t, H'_t)} = \widehat{D}_{\mathrm{KL}}^{(L)}(\sigma; H_t, H'_t)$, whose expectation over $\sigma \sim P_\theta^{(L)}(\cdot \mid u_t, H_t)$ gives the first equality. Dividing Equation~\ref{eq:app_full_factorization} by Equation~\ref{eq:app_window_factorization} gives the conditional tail distribution:
\begin{equation}
    P_\theta(\tau_{>t+L} \mid u_t, [c, \sigma]) = \frac{P_\epsilon(s_{t+1:T} \mid u_t, a_{t:T-1})}{P_\epsilon(s_{t+1:t+L} \mid u_t, a_{t:t+L-1})} \cdot \prod_{i=t+L}^{T} \pi_\theta\big(a_i \mid [c, \sigma, \tau_{t+L:i}]\big).
    \label{eq:app_tail_dist}
\end{equation}
By the sufficiency condition, every policy factor in Equation~\ref{eq:app_tail_dist} agrees across $c \in \{H_t, H'_t\}$, giving $P_\theta(\cdot \mid u_t, [H_t, \sigma]) = P_\theta(\cdot \mid u_t, [H'_t, \sigma])$ for every $\sigma$. Since $P_\theta(\tau_{>t} \mid u_t, c)$ can be expanded as $P_\theta^{(L)}(\sigma \mid u_t, c)\,P_\theta(\tau_{>t+L} \mid u_t, [c, \sigma])$, the full-trajectory KL divergence can be rewritten as:
\begin{equation}
    \begin{aligned}
        &D_{\mathrm{KL}}\big(P_\theta(\cdot \mid u_t, H_t)\,\Vert\,P_\theta(\cdot \mid u_t, H'_t)\big) \\
        &\quad = \mathbb{E}_{\tau_{>t} \sim P_\theta(\cdot \mid u_t, H_t)}\Bigg[\log\frac{P_\theta^{(L)}(\sigma \mid u_t, H_t)}{P_\theta^{(L)}(\sigma \mid u_t, H'_t)} + \log\frac{P_\theta(\tau_{>t+L} \mid u_t, [H_t, \sigma])}{P_\theta(\tau_{>t+L} \mid u_t, [H'_t, \sigma])}\Bigg] \\
        &\quad = D_{\mathrm{KL}}\big(P_\theta^{(L)}(\cdot \mid u_t, H_t)\,\Vert\,P_\theta^{(L)}(\cdot \mid u_t, H'_t)\big) \\
        &\qquad + \mathbb{E}_{\sigma \sim P_\theta^{(L)}(\cdot \mid u_t, H_t)}\Big[D_{\mathrm{KL}}\big(P_\theta(\cdot \mid u_t, [H_t, \sigma])\,\Vert\,P_\theta(\cdot \mid u_t, [H'_t, \sigma])\big)\Big].
    \end{aligned}
    \label{eq:app_kl_chain}
\end{equation}
Note that the last term in Equation~\ref{eq:app_kl_chain} is zero, which implies the second equality.
\end{proof}

\section{Description of Environments}
\label{app:environment_description}

All environments are accessed through AgentGym, which serves each benchmark as a separate HTTP server and provides a unified interface for training and evaluation~\citep{xi2025agentgym}.

\noindent \textbf{ALFWorld.} ALFWorld is a text-based embodied household environment in which the agent completes goal-oriented instructions in simulated home scenes~\citep{shridhar2021alfworld}. The instruction at each step lists the currently admissible actions. The training split contains 3,553 tasks, and the held-out test split contains 140 tasks. Each episode allows up to 50 environment steps. The reward is $1$ for a completed task and $0$ otherwise.

\noindent \textbf{WebShop.} WebShop is a simulated e-commerce website in which the agent searches for products, inspects product pages, and purchases an item that satisfies a user instruction~\citep{yao2022webshop}. Instructions with indices 500 to 6,909 form the training split of 6,410 tasks, and the first 500 instructions form the held-out test split. Each episode allows up to 20 environment steps. After each purchase, the environment computes a task score in $[0,1]$ that measures how well the purchased item matches the instruction. The training reward is $1$ when the score equals $1$ and $0$ otherwise.

\noindent \textbf{SciWorld.} SciWorld is a text-based environment for scientific experimentation that requires procedural reasoning over simulated objects and devices~\citep{wang2022scienceworld}. From the AgentGym training and test splits, we keep 14 task types grouped into five categories, which yields 1,768 training tasks and 175 test tasks. Each episode allows up to 50 environment steps. The reward is the final task score normalized to $[0,1]$, allocated only when the episode terminates.

\section{Compaction Prompt Templates}
\label{app:compaction_prompt_templates}

\begin{figure}[ht]
\begin{tcolorbox}[title=Compaction Prompt, width=\textwidth]
\begin{Verbatim}[breaklines=true, breakanywhere=true, breaksymbolleft={}, formatcom=\normalfont, fontsize=\small]
Your context is about to exceed its limit, so the interaction history above will now be removed. Write a summary to replace it. From the next step on, you will only see the task description, this summary, and the latest observation. In general, you should write the following five sections in order, and skip any section the history does not support:

1. Objective: What the task is really asking for considering the intent behind it. Restate it as you understand it rather than copying its wording.
2. History: The effective actions you have taken and what the environment answered. Reproduce important identifiers and values exactly rather than paraphrasing them.
3. Pitfalls: Mistakes the history already made and traps you can see coming. You should avoid them in the future.
4. Progress: The high-level subgoals already complete, and any partially done one together with what still stands in its way.
5. Plan: The subgoals that remain and the direction the next actions should take.

Keep the summary clean and concise. An over-long summary is truncated, keeping only its beginning and its end. First think step by step about what future actions will depend on. Then provide your summary enclosed within <summary>...</summary> tags.
\end{Verbatim}
\end{tcolorbox}
\begin{tcolorbox}[title=Resumption Prompt, width=\textwidth]
\begin{Verbatim}[breaklines=true, breakanywhere=true, breaksymbolleft={}, formatcom=\normalfont, fontsize=\small]
The earlier interaction history has been summarized due to insufficient context. You should continue working based on the summary presented as follows:
{summary}

The latest observation is presented as follows:
{observation}

{instruction}
\end{Verbatim}
\end{tcolorbox}
\caption{Prompt templates for compaction and resumption used in our experiments.}
\label{fig:compaction_prompt}
\end{figure}

All compaction methods share the compaction trigger $B_{\mathrm{trigger}}$. When compaction is triggered, the compaction prompt $P_{\mathrm{compact}}$ replaces the latest observation $s_t$ in the model input, and the summary $z$ is extracted from the \texttt{<summary>} tags of the response. If the response contains no valid summary, a fallback that reproduces the last three turns is used instead. The reconstructed context consists of the system prompt with the task description, followed by a user message that presents $z$, the latest observation $s_t$, and the action instruction. For SuffiCE, the reference window $\sigma$ is appended verbatim after this message to form $H'_{t+L}$. SuffiCE uses a compaction prompt with five sections, shown in Figure~\ref{fig:compaction_prompt} together with the resumption prompt.

\section{Detailed Budget Settings}
\label{app:budget_settings}

\begin{table}[ht]
    \caption{Budget parameters for each backbone on the ALFWorld, WebShop, and SciWorld benchmarks.}
    \label{tab:budget_parameters}
    \renewcommand{\arraystretch}{1.0}
    \centering \footnotesize \setlength{\tabcolsep}{6pt}
    \begin{tabular}{lccc}
        \toprule
            \textbf{Budget} & \textbf{AW} & \textbf{WS} & \textbf{SW} \\
        \midrule
            \multicolumn{4}{l}{\textit{Qwen3-4B}} \\
            \quad \#Turn per task & 50 & 20 & 50 \\
            \quad \#Token per generation & 4k & 4k & 4k \\
            \quad \#Token per prompt & 8k & 6k & 8k \\
            \quad \#Token per response & 8k & 6k & 8k \\
            \quad Comp. trigger $B_{\mathrm{trigger}}$ & 14k & 10k & 14k \\
            \quad Comp. threshold $B_{\mathrm{threshold}}$ & 0.5k & 0.5k & 0.5k \\
            \quad Comp. boundary $B_{\mathrm{boundary}}$ & 1k & 1k & 1k \\
        \midrule
            \multicolumn{4}{l}{\textit{Qwen2.5-7B-Instruct}} \\
            \quad \#Turn per task & 50 & 20 & 50 \\
            \quad \#Token per generation & 4k & 3k & 4k \\
            \quad \#Token per prompt & 4k & 3k & 4k \\
            \quad \#Token per response & 4k & 3k & 4k \\
            \quad Comp. trigger $B_{\mathrm{trigger}}$ & 7k & 5k & 7k \\
            \quad Comp. threshold $B_{\mathrm{threshold}}$ & 0.5k & 0.5k & 0.5k \\
            \quad Comp. boundary $B_{\mathrm{boundary}}$ & 1k & 1k & 1k \\
        \bottomrule
    \end{tabular}
\end{table}

Table~\ref{tab:budget_parameters} lists the detailed budget settings on each backbone and environment. Both backbones share the same turn limits as well as the same summary length budget boundaries. For Qwen3-4B, the token budgets for prompt, response, and trigger are twice those for Qwen2.5-7B-Instruct.

\section{Detailed Experimental Settings}
\label{app:experimental_settings}

\begin{table}[ht]
    \caption{Training and evaluation parameters of SuffiCE.}
    \label{tab:training_parameters}
    \renewcommand{\arraystretch}{1.0}
    \centering \footnotesize \setlength{\tabcolsep}{6pt}
    \begin{tabular}{lc}
        \toprule
            \textbf{Parameter} & \textbf{Setting} \\
        \midrule
            Number of training steps & 150 \\
            Tasks per batch & 16 \\
            Rollouts per task $N$ & 8 \\
            Optimizer & AdamW \\
            Learning rate & 1e-6 \\
            Weight decay & 0.01 \\
            Mini batch size & 1024 \\
            Clipping $\epsilon_{\mathrm{low}}$ & 0.2 \\
            Clipping $\epsilon_{\mathrm{high}}$ & 0.28 \\
            Gradient clipping & 1.0 \\
            Policy KL penalty & None \\
            Format error penalty & 1.0 \\
            Entropy coefficient & 0 \\
            Training temperature & 1.0 \\
            Training top-p & 1.0 \\
            Training top-k & $-1$ \\
            Evaluation temperature & 0.7 \\
            Evaluation top-p & 0.8 \\
            Evaluation top-k & 20 \\
            Maximum compactions per episode & 5 \\
            Reference window length $L$ & 3 \\
            Compatibility advantage weight $\lambda_{\mathrm{suffice}}$ & 1.0 \\
            Length budget advantage weight $\lambda_{\mathrm{budget}}$ & 1.0 \\
            Compatibility advantage clipping $\kappa$ & 3.0 \\
        \bottomrule
    \end{tabular}
\end{table}

We build on the rLLM framework with verl as the training backend~\citep{rllm2025,sheng2025hybridflow}. Table~\ref{tab:training_parameters} summarizes the detailed training and evaluation parameter settings. The clipped surrogate uses asymmetric clipping bounds as in DAPO~\citep{yu2025dapo}. All methods share the optimizer, batch size, and sampling parameters.

\section{Comparison of Learning Curves}
\label{app:learning_curve_comparison}

\begin{figure}[!ht]
    \centering
    \includegraphics[width=0.6\linewidth]{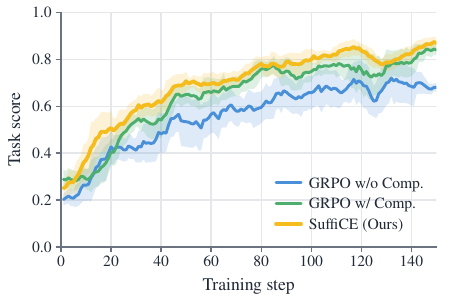}
    \caption{Training curves on SciWorld with Qwen2.5-7B-Instruct, averaged over a moving window.}
    \label{fig:training_score}
\end{figure}

\begin{figure}[ht]
    \centering
    \includegraphics[width=0.6\linewidth]{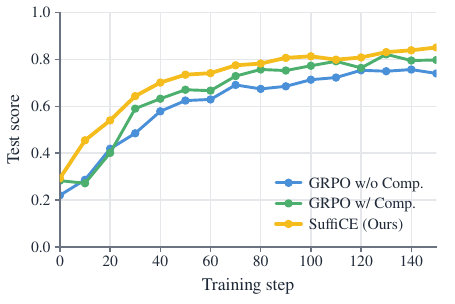}
    \caption{Evaluation curves on SciWorld with Qwen2.5-7B-Instruct, averaged over a moving window.}
    \label{fig:evaluation_score}
\end{figure}

Figures~\ref{fig:training_score} and~\ref{fig:evaluation_score} show the training and evaluation learning curves on SciWorld with Qwen2.5-7B-Instruct. On the training trajectories, SuffiCE improves faster than both non-compaction and compaction baselines. The same ordering holds on evaluation, where SuffiCE outperforms the compaction baseline and opens a large gap early in training.

\section{Ablation on Method Design}
\label{app:method_design_ablation}

\begin{table}[ht]
    \caption{Results of the method design ablation on SciWorld with Qwen2.5-7B-Instruct. We report the average task score and number of steps, compactions, and tokens in summaries. w/ $L=1$ and w/ $L=5$ vary the reference window length, w/o Scoring sets $\lambda_{\mathrm{suffice}}=0$, and w/o Budget sets $\lambda_{\mathrm{budget}}=0$.}
    \label{tab:design_ablations}
    \renewcommand{\arraystretch}{1.0}
    \centering \footnotesize \setlength{\tabcolsep}{8pt}
    \begin{tabular}{lcccc}
        \toprule
        \textbf{Method} & \textbf{\%Score} & \textbf{\#Step} & \textbf{\#Comp.} & \textbf{\#Token} \\
        \midrule
        GRPO w/ Comp. & $81.5_{\pm 0.1}$ & $21.7$ & $0.8$ & $60.1$ \\
        \midrule
        SuffiCE (Ours) & \hlfirst{$85.1_{\pm 0.5}$} & $22.2$ & $0.5$ & $157.1$ \\
        \quad w/ $L=1$ & $80.6_{\pm 1.0}$ & $20.4$ & $0.5$ & $158.6$ \\
        \quad w/ $L=5$ & \hlsecond{$84.2_{\pm 0.5}$} & $21.0$ & $0.3$ & $151.1$ \\
        \quad w/o Scoring & $80.7_{\pm 0.3}$ & $22.7$ & $0.6$ & $101.9$ \\
        \quad w/o Budget & $80.1_{\pm 0.7}$ & $32.4$ & $1.3$ & $329.2$ \\
        \bottomrule
    \end{tabular}
\end{table}

Table~\ref{tab:design_ablations} presents the method design ablation results on SciWorld with Qwen2.5-7B-Instruct. Compared with $L=3$, an overly short $L=1$ reduces the accuracy of reward estimation and weakens local supervision, whereas an overly long $L=5$ increases context pressure and causes unexpected truncation on some samples. Moreover, removing compatibility rewards causes SuffiCE to degenerate toward the compaction baseline, while removing the length budget reward induces overly long summaries. Together, these results validate SuffiCE's design.

\section{Breakdown of Time Consumption}
\label{app:time_consumption_breakdown}

\begin{figure}[ht]
    \centering
    \includegraphics[width=0.6\linewidth]{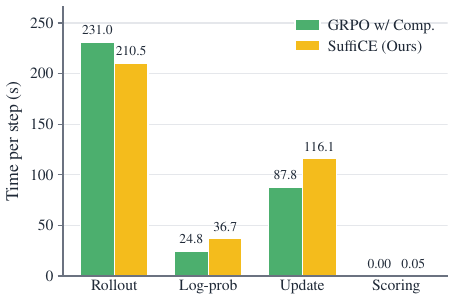}
    \caption{Per-stage wall-clock time on SciWorld with Qwen2.5-7B-Instruct for SuffiCE and the compaction baseline. For SuffiCE, Log-prob additionally involves scoring the virtual trajectories.}
    \label{fig:training_time_stage}
\end{figure}

Figure~\ref{fig:training_time_stage} compares the per-step wall-clock time of SuffiCE and the compaction baseline on SciWorld with Qwen2.5-7B-Instruct, broken down into the rollout, log-probability, update, and scoring stages. Relative to the compaction baseline, SuffiCE takes less time in rollout, as its trajectories require fewer environment steps on average. In contrast, SuffiCE incurs additional overhead in the log-probability and update stages, because the log-probability pass also evaluates the virtual trajectories $[H'_t,\sigma]$, and training each summary in a separate sequence re-encodes the preceding context. Meanwhile, reward computation in the scoring stage is negligible, taking only 0.05 seconds per step. Consequently, the rollout savings partially offset the extra training cost, keeping the overall per-step time comparable.

\section{Dynamics of Context Length}
\label{app:context_length_dynamics}

\begin{figure}[!ht]
    \centering
    \includegraphics[width=0.6\linewidth]{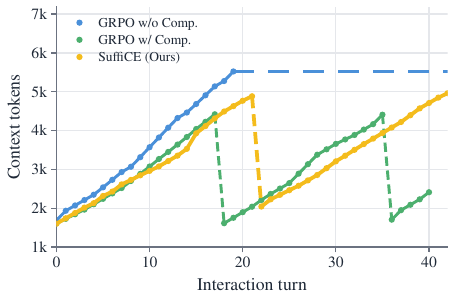}
    \caption{Turn-by-turn working context length of a representative episode. Descending dashed segments mark applied compactions, and the horizontal dashed line marks early termination.}
    \label{fig:context_length}
\end{figure}

\begin{figure}[ht]
    \centering
    \includegraphics[width=0.6\linewidth]{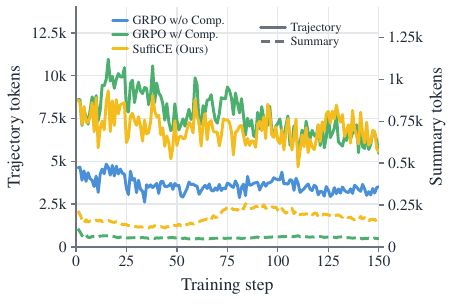}
    \caption{Trajectory and summary tokens during training for SuffiCE and the baselines. Solid curves show trajectory tokens on the left axis, and dashed curves show summary tokens on the right axis.}
    \label{fig:trajectory_length}
\end{figure}

Figure~\ref{fig:context_length} traces the turn-by-turn working context $H_t$ over a representative episode on SciWorld with Qwen2.5-7B-Instruct for SuffiCE and both non-compaction and compaction baselines. Without compaction, the context grows monotonically until exhausting the budget and terminating the episode early. In contrast, compaction methods reset the context upon summarization, allowing the agent to sustain a longer interaction horizon. Relative to the compaction baseline, SuffiCE covers a comparable horizon with only a single compaction instead of two, while resuming from a slightly larger rebuilt context $H'_{t+L}=[H'_t,\sigma]$ because deferred summarization retains the reference window $\sigma$ verbatim.

Figure~\ref{fig:trajectory_length} tracks trajectory length and summary length during training. Both compaction methods initially explore longer trajectories and gradually shorten them as the policies learn more efficient solutions. However, their summary lengths diverge throughout training. Under sparse outcome rewards, the compaction baseline produces increasingly terse summaries, whereas SuffiCE maintains fuller summaries under compatibility rewards, retaining key task details within the length budget.

\section{Analysis on Compaction Behavior}
\label{app:compaction_behavior}

\begin{table}[t]
    \caption{Compaction statistics across all benchmarks and backbones, pooled over three seeds. We report the average number of applied compactions, summary tokens, context tokens before and after compaction, and trajectory tokens. We group episodes by applied compactions, showing each group's episode share and corresponding task metric.}
    \label{tab:compaction_statistics}
    \renewcommand{\arraystretch}{1.0}
    \centering \footnotesize \setlength{\tabcolsep}{3pt}
    \begin{tabular}{llc|cccc|ccc|ccc}
        \toprule
            \multirow{2}{*}{\textbf{Benchmark}} & \multirow{2}{*}{\textbf{Method}} & \multirow{2}{*}{\textbf{\#Comp.}} & \multicolumn{4}{c|}{\textbf{\#Token}} & \multicolumn{3}{c|}{\textbf{\%Share}} & \multicolumn{3}{c}{\textbf{\%Metric}} \\
            & & & Sum. & Pre. & Post. & Traj. & 0 & 1 & $\geq$2 & 0 & 1 & $\geq$2 \\
        \midrule
            \multicolumn{13}{l}{\textit{Qwen3-4B}} \\
            \multirow{3}{*}{\quad ALFWorld} & Untrained w/ Comp. & 1.7 & 79 & 7.0k & 0.6k & 14.0k & 29.8 & 11.2 & 59.0 & 100.0 & 78.7 & 7.7 \\
             & GRPO w/ Comp. & 0.9 & 122 & 7.1k & 0.7k & 9.4k & 61.4 & 13.6 & 25.0 & 100.0 & 100.0 & 23.8 \\
             & SuffiCE (Ours) & 0.5 & 265 & 8.5k & 2.0k & 7.8k & 66.0 & 18.3 & 15.7 & 100.0 & 89.6 & 33.3 \\
        \cmidrule{2-13}
            \multirow{3}{*}{\quad WebShop} & Untrained w/ Comp. & 1.4 & 102 & 4.9k & 0.9k & 8.5k & 16.0 & 35.2 & 48.8 & 70.2 & 14.4 & 3.3 \\
             & GRPO w/ Comp. & 0.3 & 97 & 4.7k & 0.8k & 3.6k & 78.1 & 18.7 & 3.2 & 87.6 & 45.1 & 6.5 \\
             & SuffiCE (Ours) & 0.1 & 246 & 6.0k & 2.4k & 3.5k & 88.3 & 11.7 & 0.1 & 91.6 & 31.0 & 0.0 \\
        \cmidrule{2-13}
            \multirow{3}{*}{\quad SciWorld} & Untrained w/ Comp. & 1.3 & 67 & 7.9k & 1.7k & 12.8k & 7.6 & 57.9 & 34.5 & 24.3 & 17.3 & 12.5 \\
             & GRPO w/ Comp. & 1.1 & 90 & 8.0k & 1.7k & 11.0k & 36.8 & 23.4 & 39.8 & 79.2 & 80.0 & 62.7 \\
             & SuffiCE (Ours) & 0.5 & 258 & 8.8k & 2.7k & 8.5k & 55.8 & 37.0 & 7.2 & 84.5 & 75.5 & 63.0 \\
        \midrule
            \multicolumn{13}{l}{\textit{Qwen2.5-7B-Instruct}} \\
            \multirow{3}{*}{\quad ALFWorld} & Untrained w/ Comp. & 3.0 & 63 & 3.9k & 0.6k & 12.3k & 13.3 & 14.3 & 72.4 & 100.0 & 100.0 & 14.5 \\
             & GRPO w/ Comp. & 0.7 & 41 & 4.0k & 0.6k & 4.5k & 55.5 & 30.7 & 13.8 & 100.0 & 100.0 & 58.6 \\
             & SuffiCE (Ours) & 0.5 & 93 & 5.1k & 1.6k & 4.8k & 67.6 & 20.0 & 12.4 & 100.0 & 100.0 & 76.9 \\
        \cmidrule{2-13}
            \multirow{3}{*}{\quad WebShop} & Untrained w/ Comp. & 1.4 & 87 & 2.7k & 0.8k & 4.7k & 28.1 & 38.6 & 33.3 & 79.4 & 29.4 & 16.6 \\
             & GRPO w/ Comp. & 0.7 & 152 & 2.6k & 1.0k & 3.4k & 46.5 & 43.1 & 10.4 & 93.3 & 90.4 & 39.1 \\
             & SuffiCE (Ours) & 1.0 & 275 & 3.6k & 2.2k & 4.5k & 8.0 & 87.9 & 4.1 & 87.2 & 91.6 & 36.2 \\
        \cmidrule{2-13}
            \multirow{3}{*}{\quad SciWorld} & Untrained w/ Comp. & 1.5 & 68 & 4.8k & 1.6k & 7.7k & 34.3 & 13.3 & 52.4 & 21.4 & 26.9 & 30.6 \\
             & GRPO w/ Comp. & 0.8 & 60 & 4.8k & 1.7k & 5.6k & 40.2 & 44.0 & 15.8 & 86.3 & 83.9 & 62.4 \\
             & SuffiCE (Ours) & 0.5 & 157 & 5.6k & 2.4k & 5.6k & 64.2 & 18.7 & 17.1 & 88.1 & 88.0 & 70.6 \\
        \bottomrule
    \end{tabular}
\end{table}

\begin{figure}[!ht]
    \centering
    \includegraphics[width=0.6\linewidth]{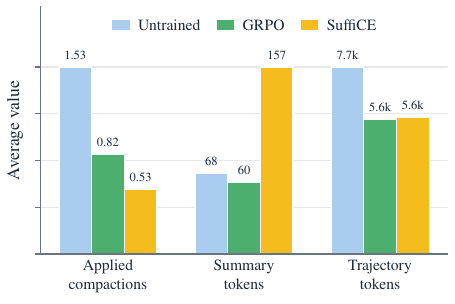}
    \caption{Compaction behavior for the untrained policy, the compaction baseline, and SuffiCE, comparing average applied compactions, summary tokens, and trajectory tokens per episode.}
    \label{fig:summary_writers}
\end{figure}

\begin{figure}[!ht]
    \centering
    \includegraphics[width=0.6\linewidth]{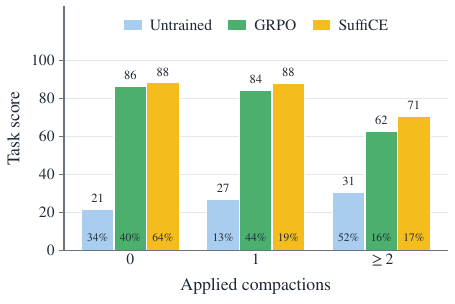}
    \caption{Task score grouped by the number of applied compactions. Values above indicate the task score, and percentages inside indicate the episode share.}
    \label{fig:score_versus_compactions}
\end{figure}

Table~\ref{tab:compaction_statistics} summarizes compaction statistics across all benchmarks and backbones, including compaction frequency, token lengths, and performance grouped by compaction count. Across all six settings, RL training substantially reduces applied compactions relative to the untrained policy, as trained agents solve tasks in fewer actions. Compared with the compaction baseline, SuffiCE generally applies fewer compactions, partly because deferred summarization avoids compacting when an episode finishes within the reference window. Meanwhile, SuffiCE generates fuller summaries and resumes from a larger rebuilt context. Consequently, on long-horizon benchmarks, SuffiCE consistently outperforms the compaction baseline on episodes requiring multiple compactions.

Figure~\ref{fig:summary_writers} further compares the compaction behavior of the untrained policy, the compaction baseline, and SuffiCE on SciWorld with Qwen2.5-7B-Instruct. Compared with the untrained policy, RL training noticeably shortens trajectories and reduces the number of applied compactions for both trained methods. Relative to the compaction baseline, whose summaries remain overly terse under sparse outcome rewards, SuffiCE generates fuller summaries within the length budget.

Figure~\ref{fig:score_versus_compactions} shows the task score breakdown by compaction count. Although performance declines on longer episodes requiring more compactions, the advantage of SuffiCE over the compaction baseline widens progressively with compaction frequency. Specifically, SuffiCE leads by 1.8 percentage points on episodes without compaction and 8.2 percentage points on those with at least two compactions, demonstrating that higher-quality summaries yield greater benefits over longer horizons.

\section{Analysis on Compaction Quality}
\label{app:compaction_quality}

\begin{table}[ht]
    \caption{N-gram recall of summaries, averaged over three seeds. History measures the share of replaced-history n-grams retained in the summary, while Future restricts this set to n-grams reused by subsequent actions.}
    \label{tab:recall_comparison}
    \renewcommand{\arraystretch}{1.0}
    \centering \footnotesize \setlength{\tabcolsep}{3pt}
    \begin{tabular}{l|ccc|ccc}
        \toprule
            \multirow{2}{*}{\textbf{Method}} & \multicolumn{3}{c|}{\textbf{History}} & \multicolumn{3}{c}{\textbf{Future}} \\
             & 1-gram & 2-gram & 3-gram & 1-gram & 2-gram & 3-gram \\
        \midrule
            Untrained w/ Comp. & $9.8$ & $4.7$ & $2.6$ & $67.0$ & $45.8$ & $28.6$ \\
            GRPO w/ Comp. & $9.6$ & $4.1$ & $2.2$ & $42.8$ & $39.1$ & $8.8$ \\
            SuffiCE (Ours) & $12.6$ & $7.2$ & $6.0$ & $64.7$ & $70.6$ & $76.9$ \\
        \bottomrule
    \end{tabular}
\end{table}

\begin{table}[ht]
    \caption{N-gram recall during training for SuffiCE and the compaction baseline. Each point averages five surrounding training steps, with Step 0 denoting trajectories from the untrained policy.}
    \label{tab:recall_trending}
    \renewcommand{\arraystretch}{1.0}
    \centering \footnotesize \setlength{\tabcolsep}{3pt}
    \begin{tabular}{ll|cccc|cccc}
        \toprule
            \multirow{2}{*}{\textbf{Method}} & \multirow{2}{*}{\textbf{N-gram}} & \multicolumn{4}{c|}{\textbf{History}} & \multicolumn{4}{c}{\textbf{Future}} \\
             &  & Step 0 & Step 50 & Step 100 & Step 150 & Step 0 & Step 50 & Step 100 & Step 150 \\
        \midrule
            \multirow{3}{*}{GRPO w/ Comp.} & 1-gram & $10.0$ & $9.6$ & $8.1$ & $7.1$ & $64.3$ & $44.3$ & $34.9$ & $52.1$ \\
             & 2-gram & $4.7$ & $3.9$ & $3.0$ & $3.4$ & $42.9$ & $42.0$ & $23.0$ & $49.3$ \\
             & 3-gram & $2.7$ & $2.3$ & $1.7$ & $2.0$ & $23.0$ & $22.7$ & $1.9$ & $21.2$ \\
        \midrule
            \multirow{3}{*}{SuffiCE (Ours)} & 1-gram & $11.6$ & $13.4$ & $19.7$ & $13.9$ & $61.9$ & $64.3$ & $74.7$ & $67.1$ \\
             & 2-gram & $5.5$ & $4.7$ & $11.2$ & $8.6$ & $44.2$ & $21.0$ & $74.3$ & $75.7$ \\
             & 3-gram & $4.1$ & $4.2$ & $9.3$ & $7.3$ & $25.4$ & $7.5$ & $75.6$ & $85.8$ \\
        \bottomrule
    \end{tabular}
\end{table}

Table~\ref{tab:recall_comparison} evaluates what generated summaries retain on SciWorld with Qwen2.5-7B-Instruct, measuring their history and future recall. Across all n-gram orders, SuffiCE preserves substantially more of the replaced history than both the untrained policy and the compaction baseline, indicating that it retains exact object names and command arguments rather than paraphrasing them away. This advantage becomes far more pronounced on decision-relevant future recall, where SuffiCE reaches 76.9\% on 3-grams compared with 28.6\% for the untrained policy and only 8.8\% for the compaction baseline.

Table~\ref{tab:recall_trending} tracks these n-gram recalls throughout training. Starting from comparable values at initialization, SuffiCE substantially improves its future 3-gram recall to 85.8\%, whereas the compaction baseline finishes near its initial level. History recall exhibits a similar divergence, declining overall for the compaction baseline while rising for SuffiCE. Together, these trends confirm that compatibility rewards provide effective local supervision for retaining decision-critical information.

\section{Case Study on SciWorld}
\label{app:sciworld_case_study}


\definecolor{swcaseenv}{RGB}{190,95,0}
\definecolor{swcaseagent}{RGB}{20,42,177}
\definecolor{swcasetag}{RGB}{100,48,155}
\definecolor{swcasepaper}{RGB}{249,250,252}
\newsavebox{\swcasebody}
\newcommand{\swcaseheader}[2]{\colorbox{#1}{\parbox{\dimexpr\linewidth-2\fboxsep\relax}{\color{white}\bfseries\footnotesize #2}}}
\newenvironment{swenvbox}[1]{%
  \def\swcasetitle{#1}\begin{lrbox}{\swcasebody}%
  \begin{minipage}{\dimexpr.9\linewidth-2\fboxsep-2\fboxrule\relax}\footnotesize\raggedright
}{%
  \end{minipage}\end{lrbox}\par\noindent
  \begin{minipage}{.9\linewidth}\swcaseheader{swcaseenv}{\swcasetitle}\par\nointerlineskip\noindent
  \fcolorbox{swcaseenv}{swcasepaper}{\usebox{\swcasebody}}\end{minipage}\par\vspace{3.5pt}
}
\newenvironment{swagentbox}[1]{%
  \def\swcasetitle{#1}\begin{lrbox}{\swcasebody}%
  \begin{minipage}{\dimexpr.9\linewidth-2\fboxsep-2\fboxrule\relax}\footnotesize\raggedright
}{%
  \end{minipage}\end{lrbox}\par\noindent\hfill
  \begin{minipage}{.9\linewidth}\swcaseheader{swcaseagent}{\swcasetitle}\par\nointerlineskip\noindent
  \fcolorbox{swcaseagent}{swcasepaper}{\usebox{\swcasebody}}\end{minipage}\par\vspace{3.5pt}
}
\newcommand{\swcaseaction}[1]{\textcolor{swcasetag}{\textless action\textgreater}#1\textcolor{swcasetag}{\textless/action\textgreater}}
\newcommand{\swcaseomit}[1]{\par\noindent\makebox[\linewidth][c]{{\footnotesize\itshape[#1]}}\par\vspace{2.5pt}}

\begin{figure}[!ht]
\setlength{\fboxsep}{3.5pt}%
\begin{swenvbox}{Environment (Step 1)}
Task: Your task is to find a(n) living thing. First, focus on the thing. Then, move it to the green box in the kitchen.\par
This room is called the living room. In it, you see: \ldots{} You also see: A door to the hallway (that is closed). [Other room details omitted.]
\end{swenvbox}
\begin{swagentbox}{Agent (Step 1)}
\swcaseaction{open(\textquotedbl door to hallway\textquotedbl)}
\end{swagentbox}
\swcaseomit{Steps 2--5: navigation to the kitchen omitted.}
\begin{swenvbox}{Environment (Step 6)}
This room is called the kitchen. In it, you see: \ldots{} a green box (containing nothing) \ldots{} [Other objects omitted.]
\end{swenvbox}
\swcaseomit{Steps 7--21: search, greenhouse entry, and plant inspection omitted.}
\begin{swenvbox}{Environment (Step 22)}
a pea plant in the reproducing stage with a tall height. On the pea plant you see: a purple flower.
\end{swenvbox}
\begin{swagentbox}{Agent (Step 22)}
\swcaseaction{focus(\textquotedbl living thing in flower pot 2\textquotedbl)}
\end{swagentbox}
\begin{swenvbox}{Environment (Step 23)}
You focus on the pea plant.
\end{swenvbox}
\begin{swagentbox}{Agent (Step 23)}
\swcaseaction{pickup(\textquotedbl living thing in flower pot 2\textquotedbl)}
\end{swagentbox}
\begin{swenvbox}{Environment (Step 24)}
You move the pea plant to the inventory.
\end{swenvbox}
\begin{swenvbox}{Compaction (Step 24)}
Your context is about to exceed its limit, so the interaction history above will now be removed. Write a summary to replace it. From the next step on, you will only see the task description, this summary, and the latest observation.
\end{swenvbox}
\begin{swagentbox}{Agent (Step 24)}
\textcolor{swcasetag}{\textless summary\textgreater}\par
\textbf{1. Objective:} The task is to find a living thing and move it to the green box in the kitchen. The goal is to locate a living organism, such as a plant or animal, and then transport it to the designated location.\par
\textbf{2. History:} I have explored multiple rooms including the living room, hallway, kitchen, and greenhouse. I found a pea plant in the greenhouse, focused on it, and picked it up. Now I need to move it to the green box in the kitchen.\par
\textbf{3. Pitfalls:} I must ensure that the pea plant is indeed a living thing and that I have the correct path to the green box. I should also avoid missing any steps in the movement process.\par
\textbf{4. Progress:} I have identified and focused on a living thing (the pea plant), and I have picked it up. The next step is to move it to the green box in the kitchen.\par
\textbf{5. Plan:} I need to go back to the kitchen, find the green box, and use the move() function to place the pea plant into it.\par
\textcolor{swcasetag}{\textless/summary\textgreater}
\end{swagentbox}
\caption{Case study from a SuffiCE evaluation trajectory, illustrating selected steps before compaction, the compaction prompt, and the summary. Environment observations and the compaction prompt are aligned on the left, while agent actions and the summary are aligned on the right.}
\label{fig:sciworld_case}
\end{figure}

Figure~\ref{fig:sciworld_case} presents a case study from a SuffiCE evaluation trajectory on SciWorld with Qwen3-4B, showing key pre-compaction steps along with the compaction prompt and generated summary~\citep{wang2022scienceworld}. Rather than retaining verbose room descriptions and intermediate navigation steps, the policy produces a concise, well-structured summary that accurately records its progress, including identifying, focusing on, and picking up the pea plant. Moreover, it preserves the exact target entities, locations, and next command needed for task completion, namely returning to the kitchen and placing the pea plant into the green box with the \texttt{move()} action. This example qualitatively confirms that SuffiCE retains decision-critical details while discarding redundant history.

\end{document}